\documentclass[11pt]{article}

\usepackage[numbers]{natbib}
\usepackage[a4paper,margin=1in]{geometry}
\usepackage[utf8]{inputenc}
\usepackage[T1]{fontenc}
\usepackage{hyperref}
\usepackage{url}
\usepackage{booktabs}
\usepackage{tabularx}
\usepackage{multirow}
\usepackage{longtable}
\usepackage{pdflscape}
\usepackage{float}
\usepackage{amsfonts}
\usepackage{amsmath}
\usepackage{amsthm}
\usepackage{amssymb}
\usepackage{nicefrac}
\usepackage{makecell}
\usepackage{siunitx}
\usepackage{csquotes}
\usepackage[final,protrusion=false,expansion=false]{microtype}
\usepackage{xcolor}
\usepackage{colortbl}
\usepackage{listings}
\lstdefinestyle{promptstyle}{
  basicstyle=\ttfamily\footnotesize,
  breaklines=true,
  breakatwhitespace=true,
  columns=fullflexible,
  keepspaces=true,
  backgroundcolor=\color{black!2},
  frame=single,
  framerule=0pt,
  framesep=4pt,
  xleftmargin=4pt,
  xrightmargin=4pt,
  showstringspaces=false,
  upquote=true,
  literate={—}{{---}}1 {×}{{$\times$}}1
}

\definecolor{diffAdd}{HTML}{1A7F37}   
\definecolor{diffDel}{HTML}{CF222E}   
\definecolor{diffHunk}{HTML}{8250DF}  
\lstdefinestyle{promptdiff}{style=promptstyle,
  morecomment=[f][\color{diffAdd}]{+},
  morecomment=[f][\color{diffDel}]{-},
  morecomment=[f][\color{diffHunk}\bfseries]{@},
}
\usepackage{graphicx}
\usepackage{subcaption}
\usepackage[affil-sl]{authblk}
\usepackage{tikz}
\usetikzlibrary{arrows.meta,positioning,shapes.geometric,calc,decorations.pathreplacing,backgrounds,fit}

\graphicspath{%
  {./figures/}%
  {./}%
}

\newcommand{\EElenchos}{E_{\mathrm{ELX}}}

\theoremstyle{plain}

\newtheorem{definition}{Definition}

\usepackage{enumitem}
\setlist{itemsep=0mm, parsep=0mm}

\begin{document}\sloppy

\title{\bf LLMs Can See the Smoke but not the Fire:\\ Evaluating Abductive Reasoning with Elenchos}

\author[2,1]{Julius Steiglechner\thanks{These authors contributed equally.}}
\author[1]{Lucas Mahler\protect\footnotemark[1]}
\author[1]{Gabriele Lohmann\thanks{Corresponding author: lohmann@tuebingen.mpg.de}}

\affil[1]{Max-Planck-Institute for Biological Cybernetics,
Magnetic Resonance Center, T{\"u}bingen, Germany}
\affil[2]{Dept. of Biomedical Magnetic Resonance, University Hospital T{\"u}bingen, T{\"u}bingen, Germany}
\date{}

\maketitle
\begin{abstract}
Large language models (LLMs) excel at pattern recognition and text generation, but their capacity for abductive inference -- inferring latent hypotheses that explain observed behavior -- remains poorly understood.
Here, we introduce {\em Elenchos} (named after the Socratic method of cross-examination), a generative evaluation framework that measures abductive reasoning as a structural inverse problem.
Given a reference formal system, such as the $\lambda$-calculus, and a potentially mutated counterpart, agents must determine whether a mutation has occurred and infer the rule modifications responsible for the resulting behavioral differences.

Evaluating frontier and mid-tier LLMs reveals a consistent detection--attribution dissociation: models often recognize that a system has been altered but struggle to identify the latent mutations causing the observed discrepancies. Performance degrades substantially under interacting mutations, where models frequently recover only a subset of the underlying mutations. Preliminary evidence also suggests diminishing returns from increased inference-time reasoning, with only modest improvements under larger reasoning budgets, though this finding requires further validation.
\end{abstract}

\section{Introduction}

Large language models excel at forward reasoning: predicting sequences, deriving consequences, or generating outputs from known rules. However, many real-world cognitive tasks -- from debugging software and diagnosing system failures to generating scientific hypotheses -- require the inverse operation: inferring hidden causes from observed behavior. We refer to this capability as abductive reasoning.

To measure this ability, we introduce Elenchos, an evaluation framework for abductive resoning over rule modifications under black-box access. The task is formulated as an inference problem over a finite hypothesis class of compositional mutations applied to an underlying rule system. Agents are presented with a reference kernel and a potentially mutated counterpart. Under a limited query budget, they must use observed behavioral differences to infer the latent rule changes responsible for any discrepancies.
Here, an \enquote{agent} denotes the entity performing inference within the Elenchos framework, such as an LLM, a human participant, or another computational system.
Unlike program analysis settings that grant access to internal source code, Elenchos restricts reasoning to a curated ontology of possible mutations under pure black-box access, enabling tightly controlled scaling of diagnostic difficulty.

Our evaluations of frontier and mid-tier LLMs using this framework reveal a consistent detection--attribution dissociation. While most models easily detect that a system has been modified, they fail to recover the latent mutations responsible for the observed behavioral changes. Metaphorically, models can see the smoke but struggle to find the fire. This gap widens when multiple mutations interact nonlinearly. 
Preliminary observations further suggest that this limitation may persist even with additional inference-time compute, hinting that abductive reasoning constitutes a distinct challenge for current reasoning systems.

Elenchos is also designed to address a common limitation of static evaluation suites: instance reuse and implicit adaptation~\citep{plateau2026}.
Because evaluation instances are generated procedurally through compositional rule mutations, the space of possible problems is combinatorially large and can be expanded without altering the underlying evaluation methodology. This reduces reliance on repeated or memorized instances, ensuring evaluation remains sensitive to generalization rather than exposure to fixed test sets while preserving the identifiability of latent rule modifications.

We instantiate Elenchos using a dependently typed $\lambda$-calculus kernel~\citep{CIC,MartinLof1984-ITT,Church1932}. The framework is defined at the level of an abstract evaluation procedure and, in principle, can be instantiated in a range of alternative computational substrates, e.g., the SKI combinator calculus~\citep{SKI1,SKI3}. This may help reduce reliance on representation-specific heuristics and probe whether performance generalizes beyond $\lambda$-calculus-structured inputs.

\section{Related Work}\label{sec:related}

Elenchos draws on four strands of literature: (i)~the saturation of static reasoning benchmarks, (ii)~LLMs as agents over formal systems, (iii)~differential, mutation, and metamorphic testing, and (iv)~fault localization and abductive inference. It recombines these around a single structural inversion: the formal system itself is treated as the object of diagnosis rather than a trusted oracle.

\paragraph{Benchmark saturation and reasoning robustness.}
Static reasoning benchmarks spanning mathematics~\citep{Hendrycks2021,gsm8k,FrontierMath}, graduate-level knowledge~\citep{rein2023gpqa,HLE,xHLE}, and broad capability suites~\citep{MMLU,gaia,bigbench} are increasingly saturated for frontier models~\citep{plateau2026,AIIndex2026}. This has raised concerns that reported gains partly reflect contamination or memorization rather than invariant reasoning ability~\citep{Bender2021,McIntosh2026,Chen2026,Tenenbaum2026}. Two complementary responses are particularly relevant: dynamic benchmark generation, which replaces static datasets with procedurally defined distributions to reduce leakage~\citep{white2024livebench}, and robustness analyses showing sharp performance degradation under semantics-preserving perturbations. For instance, GSM-Symbolic reports up to a $65\%$ drop in accuracy under prompt perturbations~\citep{mirzadeh2024gsmsymbolic}. Elenchos adopts a similar dynamic philosophy but shifts the locus of perturbation from inputs to the evaluation substrate itself, inducing a structured distribution over latent rule modifications.

\paragraph{Agents over formal systems and kernel verification.}
A substantial body of work studies LLMs as agents operating over formal systems such as proof assistants, where the goal is to construct derivations under a fixed and trusted kernel~\citep{polu2020generative,leandojo,baldur2023,deepseekprover,deepseekproverv2,alphaproof2025,miniF2F,FormalMATH2025}. In parallel, verification efforts aim to establish the correctness of such kernels themselves, e.g., via certified type checkers in MetaCoq~\citep{metacoq2020,metacoq2021tour} or Lean4Lean for Lean~\citep{lean4lean25}. Elenchos departs from both paradigms: the kernel is neither a trusted oracle nor a system to be formally verified, but a potentially corrupted object whose integrity must be inferred from behavior alone. Our reference system follows the bidirectional, normalization-by-evaluation lineage of dependent type theory implementations such as LambdaPi~\citep{lambdapy}, Altenkirch--Kaposi~\citep{AltenkirchKaposi2016}, and smalltt~\citep{smalltt2023}.

\paragraph{Mutation testing, fault localization, and abduction.}
Methodologically, Elenchos builds on differential, mutation, and metamorphic testing traditions~\citep{McKeeman1998,Csmith2011,EMI2014,Jia2011,Metamorphic}, which have recently been adapted to evaluate LLMs in software engineering settings~\citep{ubert2022,llmorpheus2024,mile2024,li2024mutationconsistency,haroon2026mutation,evosafe2025}. In contrast to these approaches, which primarily assess model robustness or debugging performance under externally introduced faults, Elenchos treats mutations as latent variables to be inferred rather than observed perturbations.

This distinction is particularly sharp in comparison to mutation-based LLM evaluation frameworks such as LLMorpheus~\citep{llmorpheus2024}, where mutated programs serve as test cases for model reasoning. In Elenchos, mutations instead operate at the level of the formal system's evaluation and typing rules, and the active mutation configuration corresponds to a latent ground-truth element drawn from a finite hypothesis class.

Unlike classical fault localization methods, which aim to identify faulty program locations or repair targets~\citep{papadakis2015metallaxis,autofl2024,llmao2024,betterDebugging2024,devlore2025}, Elenchos requires identification of abstract mutation classes from a predefined ontology. The presence of interacting mutations induces combinatorial behavior in the observable system, preventing shortcut elimination strategies and turning diagnosis into a structured form of abductive inference over a finite hypothesis space. In contrast to abductive NLP benchmarks grounded in informal narratives~\citep{bhagavatula2020abductive,inabhyd2024,abductiveSurvey2026,gear2025}, Elenchos operates in a fully executable environment, where each hypothesis can be deterministically validated against the underlying kernel, reducing reliance on human or model-based judgment.

\paragraph{Trusting trust and corrupted systems.}
The problem of reasoning about corrupted computation substrates has long been recognized in security and systems research. Thompson's "trusting trust" attack~\citep{thompson1984}, the xz/liblzma backdoor incident~\citep{xz2024}, and large-scale failures such as the CrowdStrike kernel-driver outage~\citep{crowdstrike2024} illustrate how failures in trusted computing bases may be observable long before their root causes are localized. While Elenchos is a basic research framework rather than a security system, these examples motivate the broader question of whether learned agents can perform behavioral diagnosis of underlying system integrity.

From this perspective, Elenchos should be understood as an evaluation of a core cognitive capability: the ability to infer latent structural corruption in rule-governed systems from finite interaction. Following Kerckhoffs's principle, the mutation ontology is fully public; only the active configuration is hidden during evaluation, ensuring that performance reflects diagnostic inference rather than information asymmetry.

\paragraph{Reasoning evaluation: chain-of-thought and robustness.}\label{sec:related:reasoning}
Chain-of-thought prompting~\citep{cot2022} established intermediate reasoning steps as a first-class object in LLM evaluation, and the subsequent reasoning-trace literature~\citep{tree-of-thoughts,let-verify-step-by-step} demonstrated large gains on math and symbolic tasks under structured inference-time search. Whether the resulting traces constitute genuine deduction has been actively contested: \citep{mirzadeh2024gsmsymbolic} (under perturbation), \citep{bender2021stochastic} (memorization critique), and \citep{Chen2026,Tenenbaum2026} (cognitive interpretation) argue, in different registers, that headline scores can mask shallow pattern reuse. Recent work on reasoning robustness under input perturbation~\citep{mirzadeh2024gsmsymbolic,nooppattern2024} and under software evolution~\citep{haroon2026mutation,evosafe2025} reaches a consistent conclusion: model performance is fragile under controlled changes that preserve task semantics. Our results in Section~\ref{sec:experiments} echo this finding in a different regime: the single-mutation bias and the detection--attribution gap are arguably the dependently-typed analogue of GSM-Symbolic's perturbation drop. The framing distinction is that Elenchos is procedurally generated under a fully verifiable oracle, and the perturbation is to the evaluator rather than the input  --  closing the loop between mutation testing, fault localization, and reasoning evaluation that has so far been pursued in separate communities.

\begin{figure}[ht]
\centering
\includegraphics[width=0.4\textwidth]{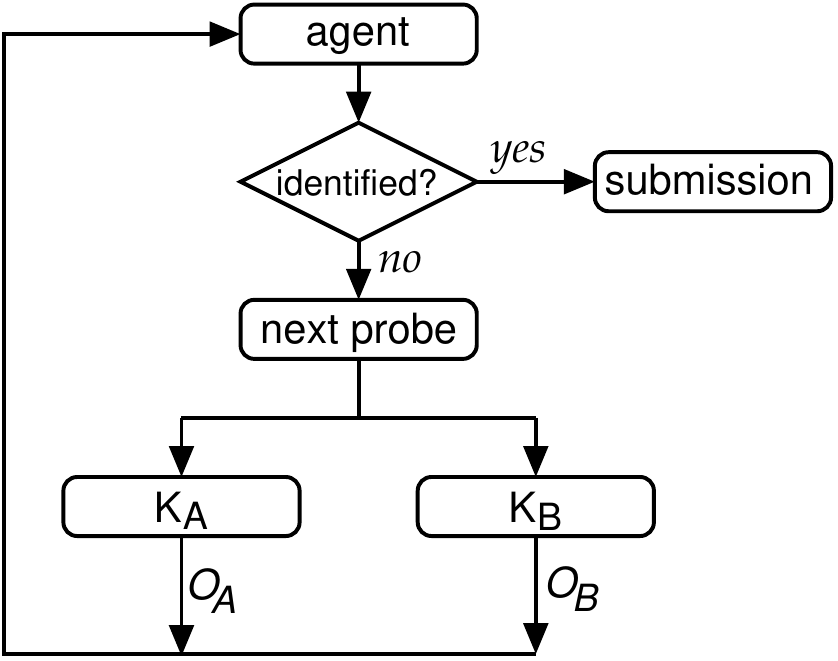}
\caption{\textbf{The Elenchos audit loop.} {\em At each iteration, the agent submits a probe to two black-box kernels, $K_A$ and $K_B$, one of which may be corrupted. The kernels return diagnostic tuples, $O_A$ and $O_B$, that summarize how the probe was processed. These tuples constitute the agent’s observations (see Section~\ref{diagnostic} for details). Using the accumulated observations, the agent assesses whether there is sufficient evidence to identify the corrupted kernel and its active mutation set. If the evidence is insufficient, the agent issues another probe; otherwise, it outputs a hypothesis specifying the corrupted kernel and the inferred mutations.}}
\label{fig:framework_pipeline}
\end{figure}

\section{The Elenchos Framework}
\label{sec:setup}

The Elenchos framework evaluates an agent's capacity for abductive reasoning by framing system diagnosis as a structural inverse problem. It consists of three core components: (i)~a formal system substrate defining the query space, (ii)~a mutation ontology defining the space of latent mutation configurations, and (iii)~an interactive auditing protocol through which an agent diagnoses those configurations under a limited probe budget.

To isolate this capacity under partial observability, we abstract the underlying rule-governed system independently of any particular formalism. While our primary evaluation instantiates this setup using the dependently typed $\lambda$-calculus kernel detailed in Section~\ref{subsec:lambda_calculus}, the general environment is formalized as follows:

\begin{definition}[Kernel]\label{def:kernel}
A \emph{kernel} $K$ is a deterministic black-box function
\[
K : \mathcal{T} \to \mathcal{O},
\]
where $\mathcal{T}$ denotes the space of probes containing all admissible syntactic terms, and $\mathcal{O}$ represents the space of observable outputs. Let $\mathcal{K}$ denote the set of all such kernels, including the reference kernel $K_{\textnormal{ref}}$ and its mutated variants.
\end{definition}

We assume that each kernel is induced by an underlying latent rule system that is not directly observable. Mutations operate on this latent representation and induce systematic changes in the observable behavior of the resulting kernel.
Agents, however, have access only to black-box interactions with this induced behavior, and must infer the latent structure solely through observed input--output discrepancies.

\begin{definition}[Mutation Ontology and Configurations]\label{def:mutation}
A \emph{mutation operator} $\mu$ is a transformation over kernels, $\mu : \mathcal{K} \to \mathcal{K}$, that modifies the underlying latent rule structure of a kernel. The mutation ontology is defined as the finite set of these primary operators:
\[
\mathcal{M} = \{\mu_1, \ldots, \mu_n\}.
\]
For any mutation configuration set $S \subseteq \mathcal{M}$, we define $K_S \in \mathcal{K}$ as the kernel obtained by applying the mutations in $S$ to the reference kernel $K_{\textnormal{ref}}$ under a fixed composition policy.
\end{definition}

\paragraph{Composite mutations.}
Mutations may interact non-linearly, meaning the behavior of a composite configuration $K_S$ cannot generally be predicted from the isolated effects of its constituent mutations. To ensure that mutation attribution is well-defined under black-box access, we restrict the evaluation space to a valid configuration family $\mathcal{F} \subseteq \mathcal{P}(\mathcal{M})$ satisfying a global \textit{identifiability} condition: for $S \in \mathcal{F}$, the mapping
\[
S \mapsto K_S
\]
is injective over $\mathcal{F}$, ensuring that each induced kernel uniquely determines its underlying mutation configuration in principle.

Elements of $\mathcal{P}(\mathcal{M})$ violating this identifiability condition are excluded from $\mathcal{F}$. Let $k = |S|$ denote the number of active mutations. For $k \ge 2$, excluded cases include: (i)~\emph{logical collapse}, where the induced kernel produces uniform outputs across the probe space, and (ii)~\emph{symptomatic subsumption}, where a configuration induces the same kernel as one of its proper subsets. By construction, the configuration family satisfies
\[
\emptyset \in \mathcal{F} \quad \text{and} \quad \{\mu\} \in \mathcal{F}, \quad \forall \mu \in \mathcal{M}.
\]

\paragraph{The audit loop.}
During a session, Elenchos evaluates abductive reasoning as an interactive diagnosis problem (\autoref{fig:framework_pipeline}). Each instance consists of two kernels $(K_A, K_B)$ implementing the same formal system, with the guarantee that at least one is sound. The agent is not informed which kernel, if either, is corrupted; instead, it interacts with both through a restricted query interface, submitting probes $\tau \in \mathcal{T}$ and observing outputs $(O_A, O_B)$. From these observations, the agent must determine whether a mutation exists, identify the corrupted kernel, if applicable, and attribute active mutations from the predefined ontology. The interaction terminates when the agent submits its final hypothesis, either on its own or after the probe budget $P_\text{max}$ is exhausted.

\paragraph{Evaluation metrics.}
{\em Kernel identification}~$V_K$ is correct if the predicted and true kernel labels match. {\em Mutation attribution}~$V_A$ is correct if the predicted and true mutation sets are identical, i.e., if {\em all} mutations have been correctly identified. After each session, a score $V \in \{0,1\}$ is returned with $V=1$ if both kernel identification and mutation attribution are correct, and $V=0$ otherwise.
Because exact-match attribution conflates missing mutations and spurious predictions, we additionally report conditional accuracy, Jaccard index, precision, and recall.

\paragraph{Inference efficiency score.}\label{sec:metric}
We propose an additional metric $\EElenchos \in [0,1.5]$ that rewards efficiency with a bonus multiplier. It is defined as
\begin{equation}
  \EElenchos(V,P) = V \left(1 + \frac{1}{2}\left(1 - \left(\frac{P}{P_{\max}}\right)^3\right)\right).
  \label{eq:eelx}
\end{equation}

The score assigns zero to incorrect submissions and rewards correct solutions with a bonus that depends on the number of probes used. 
The cubic form was chosen to provide only a mild penalty for exploratory probing while rewarding solutions obtained below the probe budget $P_\text{max}$.

\paragraph{The task is abductive, not deductive.}
The task is not primarily deductive, since success does not consist of proving statements within a fixed logical system, and it is not a standard supervised induction task, since no labeled training examples are provided within the evaluation. Rather, we operationalize abductive inference as the problem of recovering a latent mutation configuration $S \in \mathcal{F}$ that best explains observed behavioral discrepancies between paired kernels under black-box access. This corresponds to selecting a hypothesis from a finite configuration space based on partial observational evidence. This notion of abduction is consistent with its classical formulation in the sense of Peirce~\citep{Peirce1903}, where hypotheses are generated to explain observed phenomena rather than derived from axioms or learned from labeled instances.

\section{Implementation}\label{sec:implementation}

\paragraph{Dependently typed $\lambda$-calculus.}
\label{subsec:lambda_calculus}

Here we briefly review the core features of the dependently typed $\lambda$-calculus used in Elenchos, see also Supplementary Material~\ref{app:grammar}.
The set of terms $\mathcal{T}$ (which encompasses both expressions and types) is defined by the grammar:

\begin{equation}
\mathcal{T} ::= x \mid \texttt{Type} \mid \lambda x : A. M \mid M\,N \mid \Pi x : A. B,
\end{equation}

where $x$ denotes a variable, $\texttt{Type}$ represents the universe of types, $\lambda x : A. M$ is a function abstraction with domain type $A$, $M\,N$ is a function application, and $\Pi x : A. B$ is a dependent function type where the result type $B$ may depend on the argument $x$. 

Computation proceeds via $\beta$-reduction,

\begin{equation}
(\lambda x : A. M)\,N \longrightarrow_\beta M[x \mapsto N],
\end{equation}

with $M[x \mapsto N]$ denoting the capture-avoiding substitution of $N$ for $x$ in $M$. A term is in \emph{$\beta$-normal form} when no further reductions are possible. Operationally, a $\lambda$-calculus kernel acts as a deterministic type checker, validating terms against a typing context $\Gamma$ that tracks variable assignments.

\paragraph{Kernel architecture.}\label{diagnostic}
We instantiate Elenchos using \textsc{LambdaPy}, a Python-based implementation of a dependently typed $\lambda$-calculus kernel adapted from~\citep{lambdapy}.
The kernel supports parsing, elaboration, bidirectional type checking, and evaluation, providing a deterministic formal substrate for mutation-based diagnosis.
Kernels are equipped with a persistent state that includes the typing context and the evaluated definitions.
Details of the surface language are provided in Supplementary Material~\ref{app:grammar}.

Each kernel execution returns a structured diagnostic tuple:
(i) a success indicator specifying whether execution succeeds,
(ii) the normalized output value for successful executions (e.g., after applicable $\beta$-reductions and other definitional reductions supported by the kernel),
and (iii) diagnostic information for failed executions giving the type of error and an error message produced by \textsc{LambdaPy}.
These outputs implement the observation interface exposed to agents during the audit loop.
Three examples of Elenchos' cross-examination procedures are provided in the Supplementary Material~\ref{app:traces}.

\paragraph{Mutation framework.}
We use a curated ontology of $13$~singleton mutations (see Supplementary Material~\ref{app:mutations}, especially \autoref{tab:mutations_list}).
We build the corresponding valid configuration family $\mathcal{F}_k$ for $k \le 3$ according to Supplementary Material~\ref{app:nondegen} and identified \textit{additive}~$(+)$ and \textit{interacting}~$(\otimes)$, see Supplementary \autoref{fig:mutation_interactions}.

We evaluate agents on 13 singleton, 12 pairwise, and 3 triplet configurations. Among the pairwise configurations, 3 exhibit non-additive interactions, while the triplet configurations are purely additive, see Supplementary~\autoref{tab:mutation_sets}.

Rather than modifying kernel source code, mutations are implemented as configuration-level overrides. Each mutation corresponds to a dedicated flag within a runtime configuration object, with the default configuration representing the unmutated baseline. This design enables the dynamic composition of multiple mutations while maintaining isolation across evaluation instances. By decoupling mutation generation from the core kernel implementation, Elenchos supports scalable benchmark generation with controlled ground truth and reproducible mutation configurations.

\paragraph{Tool use.}
LLMs are permitted task-specific tool use, such as submitting probes, asking for remaining probe budget, listing current \textsc{LambdaPy} context, and submitting an answer.
But to prevent benchmark contamination and search-based solutions, web search and code execution are disallowed. Code execution could confound abductive reasoning with tool-use proficiency, automated hypothesis testing, exhaustive search, and external computational scaffolding. The resulting evaluations should therefore be interpreted as measuring unaided abductive attribution from observed behavior, rather than the upper bounds achievable by tool-augmented systems.

\paragraph{The system prompt.}
We evaluate four conditions that vary the level of procedural guidance provided in the system prompt. These conditions are ordered by increasing difficulty (Levels 0 through 3) as the amount of auxiliary information is gradually reduced. All levels expose the full mutation taxonomy (Supplementary \autoref{tab:mutations_list}) and differ solely in the operational hints provided. While Levels~0,\,1,\,and\,2 offer workflow guidance to the model, Level~3 completely omits this information.
The prompts are included in the Supplementary Material~\ref{app:prompts}.

\begin{itemize}

  \item \textbf{Level 0.} The system prompt includes a curated set of canonical example probes for each of the 13~mutations.\footnote{The list of constructed probes $\Pi_\texttt{RSA}$ and expected output of $K_\text{ref}$ and $K_S$ can be provided on request.} 
  Each probe is explicitly designed to elicit a distinct behavioral signature that isolates its corresponding mutation from all others. While the agent receives a complete diagnostic reference suite, it must still infer which specific subset of mutations is active in the current instance. Additionally, the prompt specifies the exact cardinality $k$ of the active mutation set.

  \item \textbf{Level 1.} The system prompt provides the exact cardinality $k$ of the active mutation set, but omits the canonical example probes.

  \item \textbf{Level 2.} The system prompt informs the agent that there are $k\leq 2$ active mutations, but does not provide the exact value of $k$.

  \item \textbf{Level 3.} The system prompt only informs the agent that there are at most three active mutations ($k \in \lbrace 0,1,2,3 \rbrace$), and does not provide workflow guidance or hints anymore.
\end{itemize}

\begin{figure}[t]
\centering
\includegraphics[width=0.95\textwidth]{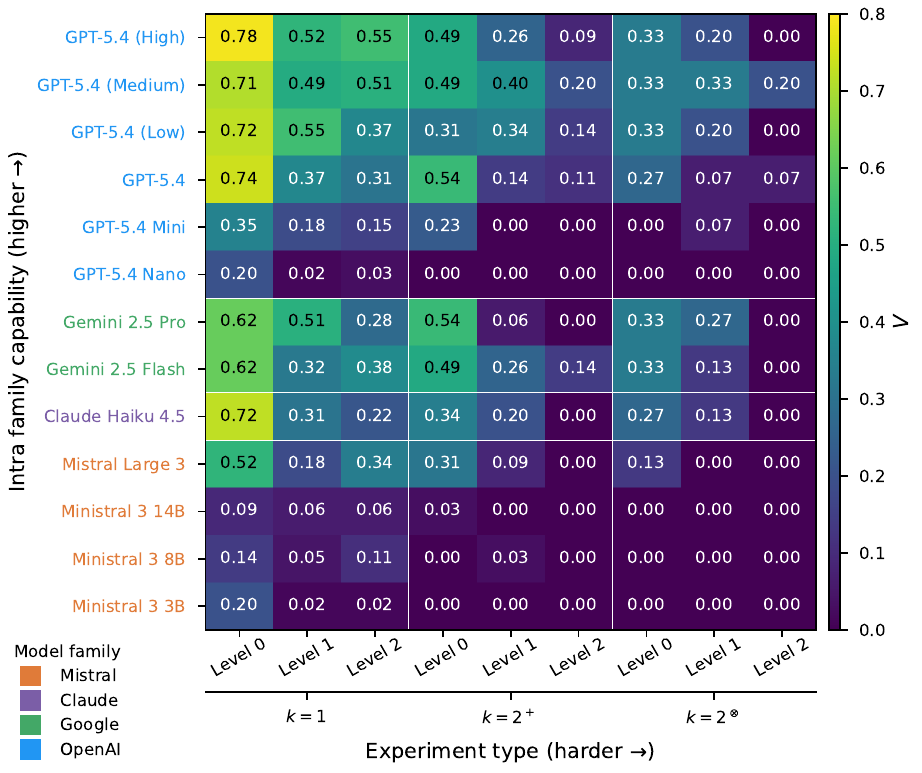}
\caption{{\bf Performance across experimental configurations (\enquote{Ladder} experiment).}
{\em Accuracy increases with model capability but decreases with reduced system prompt information (Levels), increased mutation complexity~$(k)$, and non-additive mutation interactions~$(+ \rightarrow \otimes)$.
The reported metric is the averaged exact-match accuracy~$(V)$ across mutation sets, which requires correct identification of whether either kernel is corrupted and, when applicable, the precise attribution of all active mutations.}}
\label{fig:accuracy_heatmap}
\end{figure}

\begin{table}[t]
\centering
\caption{{\bf Detection--attribution dissociation on the \enquote{Ladder} experiment, averaged over $k\in\{1,2\}$ and the three prompt levels (Levels 0--2), with models grouped by family.}
{\em  $V_K$ denotes kernel-correct detection; $V_A$ denotes exact mutation-set attribution; and $V_A \mid V_K=1$ denotes exact attribution conditioned on correct kernel detection.
Cells display mean point estimates with asymmetric Wilson $95\,\%$~confidence intervals in brackets; the Gap column reports $V_K - V_A$ with paired bootstrap $95\,\%$~confidence intervals.}}
\begin{tabular}{lrrrr}
\toprule
Model & $V_K$ & $V_A$ & $V_A \mid V_K = 1$ & Gap\,(pp) \\
\midrule
\texttt{ministral-3b-2512} & $0.33$\,{\footnotesize $[0.29, 0.38]$} & $0.11$\,{\footnotesize $[0.08, 0.14]$} & $0.13$\,{\footnotesize $[0.08, 0.20]$} & $0.23$\,{\footnotesize $[0.17, 0.28]$} \\
\texttt{ministral-8b-2512} & $0.22$\,{\footnotesize $[0.18, 0.26]$} & $0.13$\,{\footnotesize $[0.10, 0.17]$} & $0.27$\,{\footnotesize $[0.18, 0.38]$} & $0.08$\,{\footnotesize $[0.03, 0.13]$} \\
\texttt{ministral-14b-2512} & $0.31$\,{\footnotesize $[0.26, 0.36]$} & $0.05$\,{\footnotesize $[0.03, 0.08]$} & $0.14$\,{\footnotesize $[0.09, 0.22]$} & $0.26$\,{\footnotesize $[0.21, 0.30]$} \\
\texttt{mistral-large-2512} & $0.65$\,{\footnotesize $[0.59, 0.69]$} & $0.31$\,{\footnotesize $[0.27, 0.36]$} & $0.38$\,{\footnotesize $[0.32, 0.44]$} & $0.33$\,{\footnotesize $[0.27, 0.40]$} \\
\texttt{claude-haiku-4-5} & $0.78$\,{\footnotesize $[0.74, 0.82]$} & $0.34$\,{\footnotesize $[0.29, 0.39]$} & $0.39$\,{\footnotesize $[0.34, 0.45]$} & $0.44$\,{\footnotesize $[0.38, 0.50]$} \\
\texttt{gemini-2.5-flash} & $0.89$\,{\footnotesize $[0.85, 0.92]$} & $0.36$\,{\footnotesize $[0.31, 0.41]$} & $0.40$\,{\footnotesize $[0.35, 0.46]$} & $0.53$\,{\footnotesize $[0.47, 0.58]$} \\
\texttt{gemini-2.5-pro} & $0.93$\,{\footnotesize $[0.90, 0.95]$} & $0.35$\,{\footnotesize $[0.31, 0.41]$} & $0.38$\,{\footnotesize $[0.33, 0.43]$} & $0.57$\,{\footnotesize $[0.52, 0.63]$} \\
\texttt{gpt-5.4-nano} & $0.16$\,{\footnotesize $[0.13, 0.20]$} & $0.08$\,{\footnotesize $[0.05, 0.11]$} & $0.29$\,{\footnotesize $[0.18, 0.41]$} & $0.09$\,{\footnotesize $[0.05, 0.12]$} \\
\texttt{gpt-5.4-mini} & $0.58$\,{\footnotesize $[0.53, 0.63]$} & $0.23$\,{\footnotesize $[0.19, 0.28]$} & $0.27$\,{\footnotesize $[0.21, 0.34]$} & $0.35$\,{\footnotesize $[0.28, 0.41]$} \\
\texttt{gpt-5.4} & $0.86$\,{\footnotesize $[0.82, 0.90]$} & $0.37$\,{\footnotesize $[0.32, 0.42]$} & $0.42$\,{\footnotesize $[0.37, 0.48]$} & $0.50$\,{\footnotesize $[0.45, 0.55]$} \\
\texttt{gpt-5.4\_low} & $0.94$\,{\footnotesize $[0.91, 0.96]$} & $0.41$\,{\footnotesize $[0.36, 0.47]$} & $0.44$\,{\footnotesize $[0.39, 0.50]$} & $0.52$\,{\footnotesize $[0.47, 0.57]$} \\
\texttt{gpt-5.4\_medium} & $0.92$\,{\footnotesize $[0.89, 0.94]$} & $0.47$\,{\footnotesize $[0.42, 0.52]$} & $0.51$\,{\footnotesize $[0.46, 0.57]$} & $0.45$\,{\footnotesize $[0.40, 0.50]$} \\
\texttt{gpt-5.4\_high} & $0.89$\,{\footnotesize $[0.85, 0.92]$} & $0.46$\,{\footnotesize $[0.41, 0.51]$} & $0.51$\,{\footnotesize $[0.46, 0.57]$} & $0.43$\,{\footnotesize $[0.38, 0.48]$} \\
\midrule
Overall & $0.65$\,{\footnotesize $[0.64, 0.66]$} & $0.28$\,{\footnotesize $[0.27, 0.30]$} & $0.39$\,{\footnotesize $[0.37, 0.41]$} & $0.37$\,{\footnotesize $[0.35, 0.38]$} \\
\bottomrule
\end{tabular}

\label{tab:det_attr_gap}
\end{table}

\begin{figure}[t]
\centering
\includegraphics[width=0.8\textwidth]{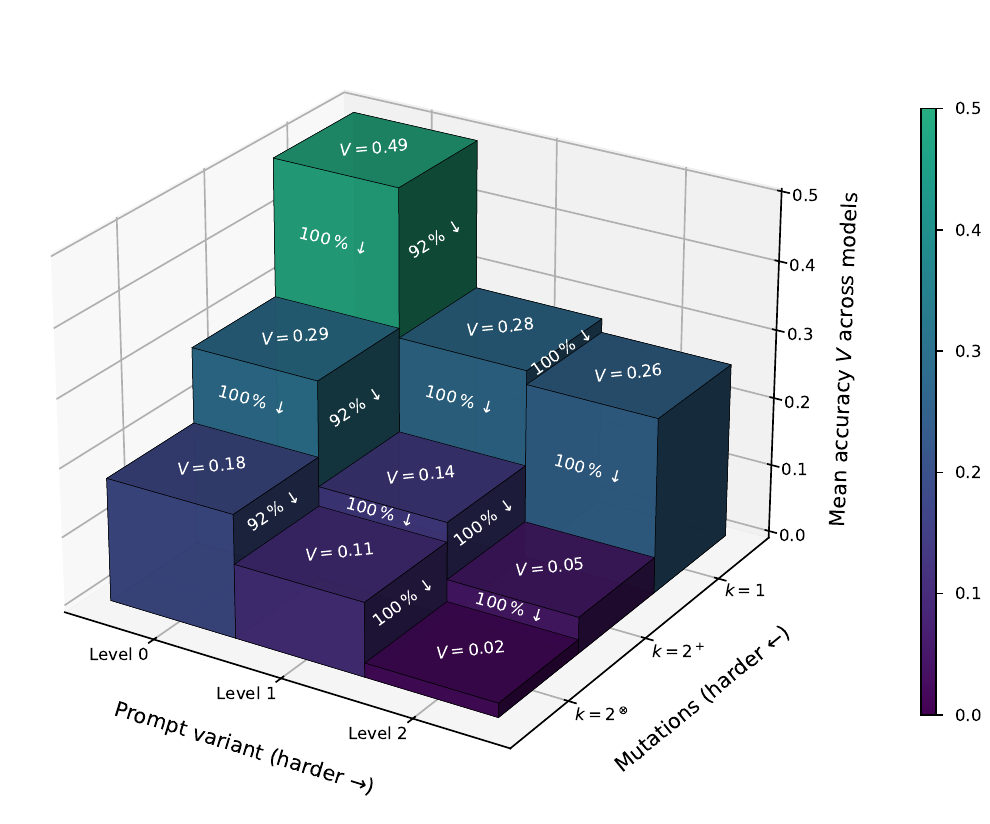}
\caption{{\bf Summary of the \enquote{Ladder} results averaged across all evaluated LLMs.} {\em Performance degrades consistently with reductions in system prompt information (Level), higher mutation counts~$(k)$, and the presence of interacting mutations~$(+ \rightarrow \otimes)$.
At the most challenging configuration (minimum prompt information combined with interacting mutations), the average accuracy drops to $V=0.02$.}}
\label{fig:difficulty_partial_order}
\end{figure}

\section{Experiments and Results}\label{sec:experiments}

We evaluate Elenchos across two distinct experimental designs:
\begin{itemize}
\item{\bf Difficulty Ladder:} 
A deep-dive evaluation using a select group of LLMs to analyze how varying levels of system prompt scaffolding together with the number of active mutations $k$ 
affect model performance.
\item{\bf Wide-Panel Benchmark:} 
A broader leaderboard evaluation that tests a diverse array of frontier and mid-tier LLMs against a fixed baseline difficulty (system prompt Level~3, and number of active mutations $k \leq 3$) 
to establish a comparative performance baseline.
\end{itemize}

\subsection{Difficulty Ladder}\label{sec:exp_ladder}

The \enquote{Ladder} experiment evaluates 13 LLMs across two mutation counts ($k \in \{1,2\}$) and three prompt levels (Levels~0,\,1,\,and\,2). We set the maximum probe budget to $P_{\max}=50$. We use 13 singleton mutations and 10 pair-wise mutations, three of which exhibit non-additive interaction effects ($k=2^\otimes$) (see Supplementary \autoref{tab:mutation_sets}). Each prompt-level and mutation configuration combination is evaluated over 5~independent repetitions.
The results are shown in \autoref{tab:det_attr_gap}, \autoref{fig:accuracy_heatmap}, and \autoref{fig:difficulty_partial_order}.
A detailed per-model perspective on the effect of the mutation set complexity and the effect of different hint levels in the system prompt is provided in the Supplementary \autoref{tab:multi_mutation_effect} and \ref{tab:hint_effect}.

\paragraph{Performance gradient.}
As difficulty increases, accuracies decrease, driven by reduced system prompt information and increasing mutation complexity (overall see \autoref{fig:difficulty_partial_order}\, per-model see \autoref{fig:accuracy_heatmap}). This gradient suggests that the difficulty scales as intended: tasks remain achievable under favorable conditions, while progressively challenging frontier models.
In the most challenging configuration (minimal prompt information with interacting mutations), average accuracy drops to $V=0.02$.

\paragraph{The detection--attribution gap.}
Averaged across all 13~LLMs and over $k\in\{1,2\}$ and prompt levels, models achieve a kernel detection accuracy of $V_K=0.65$, but an exact mutation attribution accuracy of only $V_A=0.28$, revealing a substantial detection--attribution gap (\autoref{tab:det_attr_gap} visualized in Supplementary \autoref{fig:ladder_heatmap_detail}). This discrepancy persists across all evaluated models; the largest gaps appear among the highest-performing models (e.g., \texttt{gemini-2.5-pro}), while lower-performing models also retain substantial separation (e.g., \texttt{ministral-3b-2512}).

\paragraph{System prompt effects.}
Providing more structural information through Level-0 system prompts yields substantial performance gains over Level-1 and Level-2 prompts. Weaker models fail almost completely without strong prompt scaffolding (\autoref{fig:accuracy_heatmap}).

For $k=2$ and conditioned on correct kernel detection ($V_K=1$), providing mutation-count information shifts attribution failures: models are more likely to hallucinate mutations when $k$ is known and more likely to omit mutations when it is not (Supplementary \autoref{tab:reporting}).
In other words, when $k$ is known, the model seems to feel compelled to fill all slots, leading to hallucinated mutation identities. When $k$ is not known, the model leaves mutations unreported.

\paragraph{Effects of interacting mutations.}
Performance decreases substantially when mutations exhibit non-additive interactions (\autoref{fig:accuracy_heatmap} and across prompt levels in Supplementary \autoref{tab:multi_mutation_effect}).
At $k=2$ with Level-2 prompts, most models fail completely, with only two exceptions: \texttt{gpt-5.4\_medium} with $V=0.20$ and \texttt{gpt-5.4} with $V=0.07$. Notably, the highest reasoning-budget setting achieves zero accuracy in this case. The results using a Level-1 prompt show a similar pattern: performance drops sharply from \enquote{additive} to \enquote{interacting} mutations with only two exceptions (\texttt{gpt-5.4-mini} and \texttt{gemini-2.5-pro}).

\paragraph{Per-mutation difficulty.}
Attribution difficulty is highly heterogeneous across the mutation ontology: some mutation classes are recovered by nearly all models, whereas others remain near the performance floor (Supplementary \autoref{fig:mutation_difficulty}).

\paragraph{Probe budget limitations.}
In $86.5\,\%$ of cases where models correctly detected the corrupt kernel but failed exact attribution ($V_K=1, V_A=0$), the allocated probe budget remained unused.
This suggests that many failures occur before the probe budget becomes a limiting factor, consistent with premature commitment to incorrect hypotheses rather than insufficient interaction opportunities (Supplementary \autoref{tab:reporting}).

\paragraph{Baselines comparison.}
A random mutation-selection baseline over the 13-mutation ontology achieves accuracies of $1/13 \approx 0.077$ for $k=1$ and $1/64 \approx 0.016$ for $k=2$, so most models substantially exceed chance performance. Randomly selecting the corrupted kernel among $\{A,B,\mathrm{neither}\}$ yields a $1/3$ baseline accuracy. A naive exhaustive-search baseline is computationally infeasible due to the combinatorial size of the probe space $\mathcal{T}$.

\subsection{Wide-Panel Benchmark}
\label{sec:exp_bench}

\begin{figure}[p]
\centering
\includegraphics[width=0.999\textwidth]{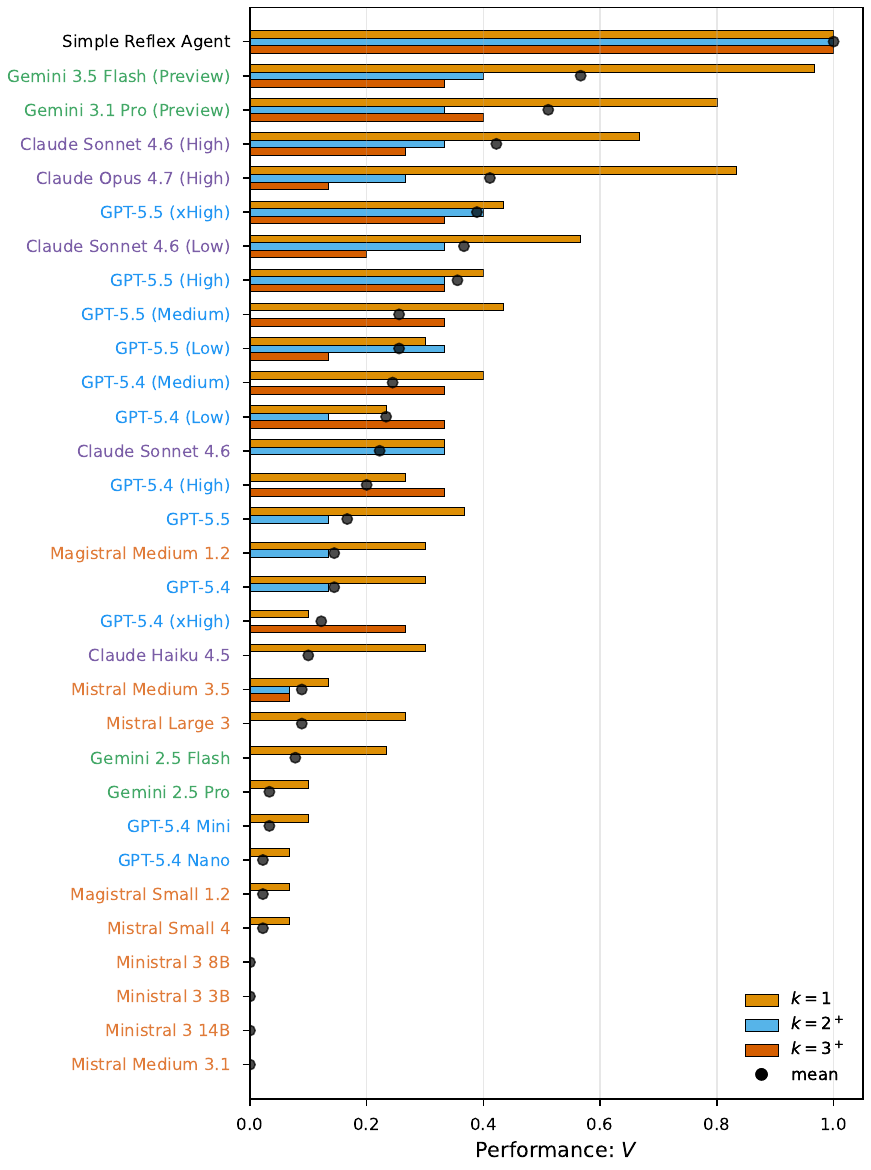}
\caption{{\bf Performance leaderboard across models and mutation orders in \enquote{Benchmark}.}
{\em For this comparison, only additive $k^+$ composite mutations have been used. $k \ge 2^\otimes$ interacting mutations have been excluded since the performance on these sets has been close to~$0$ across all models (see Supplementary \autoref{tab:full_benckmark}). The \texttt{Simple Reflex Agent} denotes a handcrafted agent that does not generate probes of its own, but simply uses the example probes specified with the Level-0 system prompt (see Section~\ref{sec:rsa}). Hence, it reaches an accuracy of~$V=1$.
}}
\label{fig:leaderboard}
\end{figure}

The wide-panel \enquote{Benchmark} evaluates 30~LLMs under a fixed baseline configuration (Level-3 system prompt) across mutation set cardinality $k \in \{0, 1, 2, 3\}$.
We use 6~singleton, 3~pair-wise additive, and 3~triplet additive mutation sets (see Supplementary \autoref{tab:mutation_sets}).
Importantly, this subset excludes interacting mutations for now, focusing strictly on purely additive configurations. This restriction was introduced because performance degraded too severely under interacting mutations to yield an informative evaluation, see Section~\ref{sec:exp_ladder}.
However, we separately tested one interacting mutation set ($k=2^\otimes$).
For each $(\text{LLM}, S)$ pair, we evaluate 5~independent sessions with a maximum probe budget of $P_{\max}=65$.
\autoref{fig:leaderboard} shows the resulting performance leaderboard, and \autoref{tab:k_ladder} shows the results averaged across all LLMs included in the leaderboard.
Supplementary \autoref{tab:full_benckmark} and \autoref{fig:leaderboard_detailed} provides a detailed evaluation.
Findings regarding the probe budget limitations are supported by Supplementary \autoref{fig:probing_behavior}.

\paragraph{The Simple Reflex Agent.}
The \texttt{Simple Reflex Agent (RSA)} is a deterministic baseline that does not generate custom probes, but instead relies exclusively on the static example probes $\Pi_\texttt{RSA}$ provided in the Level-0 system prompt. By construction, these probes, together with the expected output of reference and corrupted kernels $(O_\text{ref}, O_S)$, serve as an analytical oracle that yields perfect identification and attribution accuracy at $k=1$ and additive sets.
Crucially, none of the evaluated LLMs was capable of generating structurally similar probes autonomously to achieve $V=1$ in the \enquote{Benchmark}.
Furthermore, even when explicitly provided with these diagnostic probes via the system prompt -- as evaluated in the \enquote{Ladder} experiment -- the models failed to utilize them, underperforming significantly despite having access to an optimal query set.
For more details of constructing the oracle probes and \texttt{RSA}, see Supplementary Material~\ref{sec:rsa}.

\paragraph{Identifying the null case.}
At $k=0$, the correct kernel classification is \texttt{neither}, indicating that both kernels are sound. Across all 30~models and \texttt{RSA} and 155~experimental sessions, models achieved a null-case specificity of $0.90$. This performance provides evidence against trivial shortcut strategies, such as uniformly predicting that a system has been corrupted. The number of probes expended varies across models: some models submit \texttt{neither} very early, whereas others do so only after exhausting the available budget.
The former may indicate over-confidence, while the latter may reflect difficulty distinguishing the null case from hard-to-detect mutation configurations (Supplementary \autoref{tab:null_baseline}).

\paragraph{Scalability and the multi-mutation collapse.}
Average exact accuracy $V$ declines sharply as the number of mutations increases, dropping from $\num{0.90} \to \num{0.32} \to \num{0.15} \to \num{0.15}$ for $k=0,1,2^+,3^+$, respectively.
For the interacting set, the average accuracy is $\num{0.03}$, see \autoref{tab:k_ladder}.
There are two severe performance collapses occurring between $k=1$ and $k \ge 2$ and between additive and interacting combinations, which aligns with the single-mutation bias identified in the \enquote{Ladder} experiment.
While basic change detection remains relatively stable beyond $k=1$, causal attribution remains the primary failure mode.

This trend is further substantiated by the Jaccard similarity scores, which reach their lowest point at $k=2^\otimes$ ($V_A^J=\num{0.27}$).
This performance valley confirms that the primary driver of degradation is a localized failure in identifying specific, interacting mutations rather than an inability to detect overall system discrepancies (\autoref{tab:k_ladder}).

\begin{table}[t]
\centering
\caption{{\bf Marginal accuracy by mutation count $k$ on the wide-panel \enquote{Benchmark}.}
{\em Exact correctness $V$, kernel detection $V_K$, exact attribution $V_A$, Jaccard partial credit $V_A^{J}$, and conditional attribution correctness $V_A \mid V_K = 1$.
For $k \ge 2$, superscripts indicate whether the mutations are additive~$(+)$ or interacting~$(\otimes)$.
Values are averaged numbers across models, system prompt levels, and mutation sets with bootstrap $95\,\%$~confidence intervals in brackets.
}}
\begin{tabular}{llrrrrr}
\toprule
$k$ & $n$ & $V$ & $V_K$ & $V_A$ & $V_A^{J}$ & $V_A \mid V_K=1$ \\
\midrule
$0$ & $155$ & $0.90$\,{\footnotesize $[0.85, 0.94]$} & $0.90$\,{\footnotesize $[0.85, 0.94]$} & $0.97$\,{\footnotesize $[0.94, 0.99]$} & $0.97$\,{\footnotesize $[0.94, 0.99]$} & $1.00$\,{\footnotesize $[1.00, 1.00]$} \\
$1$ & $930$ & $0.32$\,{\footnotesize $[0.29, 0.35]$} & $0.77$\,{\footnotesize $[0.75, 0.80]$} & $0.33$\,{\footnotesize $[0.30, 0.36]$} & $0.36$\,{\footnotesize $[0.34, 0.40]$} & $0.42$\,{\footnotesize $[0.38, 0.46]$} \\
$2^+$ & $465$ & $0.15$\,{\footnotesize $[0.12, 0.18]$} & $0.63$\,{\footnotesize $[0.58, 0.67]$} & $0.15$\,{\footnotesize $[0.12, 0.18]$} & $0.28$\,{\footnotesize $[0.25, 0.32]$} & $0.24$\,{\footnotesize $[0.19, 0.29]$} \\
$3^+$ & $465$ & $0.15$\,{\footnotesize $[0.12, 0.19]$} & $0.67$\,{\footnotesize $[0.63, 0.71]$} & $0.15$\,{\footnotesize $[0.12, 0.19]$} & $0.32$\,{\footnotesize $[0.29, 0.35]$} & $0.23$\,{\footnotesize $[0.19, 0.28]$} \\
$2^\otimes$ & $155$ & $0.03$\,{\footnotesize $[0.01, 0.06]$} & $0.74$\,{\footnotesize $[0.67, 0.81]$} & $0.03$\,{\footnotesize $[0.01, 0.06]$} & $0.27$\,{\footnotesize $[0.23, 0.32]$} & $0.04$\,{\footnotesize $[0.01, 0.09]$} \\
\bottomrule
\end{tabular}

\label{tab:k_ladder}
\end{table}

\paragraph{Effects of inference-time compute.}
Supplementary \autoref{fig:reasoning_effort} displays the effect of scaling inference-time compute (or \enquote{reasoning effort}) across three models: \texttt{gpt-5.4}, \texttt{gpt-5.5}, and \texttt{claude-sonnet-4-6}. Surprisingly, we do not observe a consistent performance improvement when transitioning from low to high reasoning-effort configurations. In the most challenging setting featuring interacting mutations, all evaluated models failed entirely regardless of the allocated compute budget. However, because the sample size for this particular experiment ($k=2^\otimes$) was very small (one mutation configuration with $n=5$ per model and reasoning effort), these results remain preliminary, and further evaluation is required to substantiate this trend.

\section{Discussion}\label{sec:discussion}
We introduced Elenchos, a framework for studying abductive diagnostic reasoning in LLMs through the inference of latent rule mutations in formal systems. By framing diagnosis as a structural inverse problem, Elenchos enables the study of a fundamental reasoning capability that remains poorly understood in current AI systems.

Across two complementary evaluation regimes, our results provide several insights into the capabilities and limitations of contemporary frontier reasoning models:

\begin{itemize}
    \item \textbf{Detection--attribution dissociation:} Models consistently detect behavioral anomalies far more reliably than they isolate their underlying causes. This gap suggests that recognizing the existence of a system discrepancy is a significantly lower cognitive hurdle than constructing a complete, valid explanatory hypothesis for it.
    \item \textbf{Interacting mutations as a critical failure mode:} Non-additive configurations present an exceptional challenge across all evaluated LLMs. Performance drops sharply in these regimes, exposing a lack of robust compositional reasoning when navigating overlapping causal effects.    
    \item \textbf{Single-mutation bias:} When multiple mutations are present, models exhibit a strong tendency to terminate their diagnostic search prematurely after identifying a single plausible explanation as soon as the information about the mutation order $k$ is missing in the system prompt.
    Because attribution failures are overwhelmingly dominated by under-reporting rather than the hallucination of non-existent mutations, current systems appear to act as heuristic hypothesis generators rather than exhaustive reasoners.
    \item \textbf{High sensitivity to prompt scaffolding:} Providing structured prompt scaffolding significantly improves attribution accuracy, indicating that the underlying diagnostic capacity is often present but requires explicit environmental conditioning to be fully leveraged.
    \item \textbf{Tentative insensitivity to inference-time compute:} 
    Preliminary results suggest that moving from low- to high-reasoning-effort configurations yields no statistically significant gains in attribution accuracy. While this compute-scaling dimension requires larger-scale testing to confirm a definitive trend, early data implies that scaling search tokens alone may not resolve fundamental abductive deficits.
\end{itemize}

These findings raise a fundamental question: Do these failure modes reflect a fundamental limitation of current AI architectures, or a capability that can be mitigated through training, representation, and reasoning strategies?

\subsection{Limitations}\label{sec:limitations}

While Elenchos provides a rigorous framework for isolating abductive capacity, our evaluation is bounded by several specific constraints.

\begin{itemize}
    \item \textbf{Domain and ontology bounds:} Our evaluation is constrained to a restricted formal language and a curated 13-mutation ontology with maximum mutation counts of $k=3$. Although this constraint ensures precise control over the ground-truth causal structure, it does not fully mirror the expansive complexity of large-scale production software or open-ended adversarial environments.
    \item \textbf{Sample scaling and edge-Case performance:} Certain auxiliary dimensions warrant broader empirical scaling. Specifically, our preliminary insights regarding the limited efficacy of inference-time compute rely on a small sample size that requires deeper validation. Furthermore, the most challenging configurations, such as pairing minimal prompt context with heavily interacting mutations, require dedicated stress-testing using the next generation of frontier LLMs.
    \item \textbf{Confounding formal-system dynamics:} Success in the current benchmark requires a model to simultaneously excel at abductive attribution and symbolic formal reasoning. The pronounced detection--attribution dissociation observed here motivates the instantiation of the framework across non-symbolic, rule-governed environments to cleanly disentangle general deficits in abductive logic from localized limitations in formal semantic parsing.
    \item \textbf{Prompt sensitivity:} Our current protocol relies on fixed prompt scaffolds. The high variance in accuracy across different prompt conditions highlights a stark sensitivity to elicitation strategies, leaving open the question of how optimal prompting strategies might close the attribution gap.
\end{itemize}

\subsection{Broader Impact and Applications}\label{sec:impact}

Although Elenchos is instantiated here using a dependently typed $\lambda$-calculus kernel, the underlying capability it measures -- inferring hidden structural modifications from behavioral anomalies -- is a fundamental component of reasoning in several practical and scientific domains.

\begin{itemize}
\item{\bf Cybersecurity and forensics:}
Security analysts routinely trace unexpected system behaviors back to root causes under conditions of incomplete information, adversarial masking, and zero-day exploitation. By testing a model's ability to untangle interacting mutations, Elenchos provides a contamination-resistant environment for evaluating the rigorous forensic reasoning required for automated incident response and vulnerability analysis.

\item{\bf System assurance and autonomous agents:}
As LLM-based systems are increasingly deployed in safety-critical orchestration and monitoring workflows, their ability to perform reliable multi-fault diagnosis becomes essential. Assessing whether an agent can move beyond surface-level anomaly detection toward complete causal attribution is vital for mitigating silent failures and preventing cascading system errors in production environments.

\item{\bf Scientific discovery and hypothesis generation:}
Many scientific breakthroughs rely on modeling unobservable mechanisms from sparse experimental data. By isolating abductive reasoning from simple pattern completion, Elenchos offers a controlled testbed to evaluate how effectively frontier models can participate in automated hypothesis generation and the discovery of latent physical or logical laws.
\end{itemize}

\subsection{Future Directions}\label{sec:future}
Disentangling whether the observed detection–-attribution gap is a persistent architectural barrier or a treatable engineering challenge remains a primary objective. Addressing this will require systematic evaluation across model scales, training paradigms, and inference-time reasoning budgets, alongside the development of external scaffolding.

A further direction is to extend Elenchos beyond the dependently typed $\lambda$-calculus setting to other rule-governed systems and surface languages. Such extensions would help distinguish general limitations in abductive attribution from difficulties arising from the particular symbolic representations used in the current benchmark.

Additional directions include studying prompt sensitivity and elicitation effects, developing adaptive verification procedures, exploring automated self-correction mechanisms, and extending the mutation ontology to larger and more diverse hypothesis spaces.

\subsection*{Acknowledgements}
This work was supported by the grant DFG SPP-2041 (DFG LO1728/2-1, SCHE 658/17-1). The authors have no competing interests.

\bibliography{lit}

\clearpage

\vspace*{5cm}
\centerline{\Huge {\bf Supplementary Material}}

\clearpage

\appendix

\section{\textsc{LambdaPy} Surface Grammar}\label{app:grammar}

The surface syntax accepted by the kernel is defined by the Lark grammar reproduced below (\texttt{lambdapy/grammar.lark}). Source strings are parsed with Lark's Earley algorithm and elaborated to locally-nameless de~Bruijn terms before reaching the bidirectional type checker.

\begin{lstlisting}[style=promptstyle]
// LambdaPi surface syntax grammar (Lark Earley)
//
// Supports both Python-style and Haskell-style syntax:
//   Python: assume x : T        Haskell: assume (x :: T)
//   Python: let x : T = body    Haskell: let x = (body :: T)
//   Python: e : T               Haskell: e :: T
//   Python: Type 0              Haskell: * (same as Type 0)
//
// Each statement occupies one logical line ending with NEWLINE.

%import common.CNAME    -> NAME
%import common.INT
%import common.NEWLINE
%import common.WS_INLINE
%ignore WS_INLINE
%ignore /--[^\n]*/

start: statement+

// ---------------------------------------------------------------------------
// Statements  (each ends with NEWLINE)
// ---------------------------------------------------------------------------

// Haskell-style: assume (a :: *) (b :: *)  or  assume a :: *
// Python-style:  assume x y : T
statement: "assume" paren_binding+ NEWLINE               -> assume_multi_stmt
         | "assume" NAME ("::" | ":") term NEWLINE       -> assume_single_stmt
         | "assume" name_list ":" term NEWLINE            -> assume_stmt
         | "let"    NAME "=" term NEWLINE                 -> let_infer_stmt
         | "let"    NAME ":" term "=" term NEWLINE        -> let_stmt
         | "eval"   term NEWLINE                          -> eval_stmt
         | "check"  term NEWLINE                          -> check_stmt

// Parenthesized binding for Haskell-style assume
paren_binding: "(" NAME ("::" | ":") term ")"

name_list: NAME+

// ---------------------------------------------------------------------------
// Terms
// ---------------------------------------------------------------------------

term: lam_term
    | forall_term
    | app_term "->" term              -> arrow
    | app_term ("::" | ":") term      -> ann
    | app_term

// Lambda
lam_term: "\\" NAME+ "->" term

// Universal (forall binder supports both : and ::)
forall_term: "forall" binder+ "." term

binder: "(" NAME ("::" | ":") term ")"

// ---------------------------------------------------------------------------
// Application (left-associative)
// ---------------------------------------------------------------------------

app_term: atom+  -> app_chain

// ---------------------------------------------------------------------------
// Atoms
// ---------------------------------------------------------------------------

atom: "(" term ")"   -> paren
    | "Type" INT?    -> universe
    | "*" INT?       -> universe
    | "Nat"          -> nat
    | "Zero"         -> zero
    | "Succ"         -> succ_kw
    | "NatElim"      -> nat_elim_kw
    | "natElim"      -> nat_elim_kw
    | "Vec"          -> vec_kw
    | "Nil"          -> nil_kw
    | "Cons"         -> cons_kw
    | "VecElim"      -> vec_elim_kw
    | "vecElim"      -> vec_elim_kw
    | "Fin"          -> fin_kw
    | "FZero"        -> fzero_kw
    | "FSucc"        -> fsucc_kw
    | "FinElim"      -> fin_elim_kw
    | "finElim"      -> fin_elim_kw
    | "Eq"           -> eq_kw
    | "Refl"         -> refl_kw
    | "EqElim"       -> eq_elim_kw
    | "eqElim"       -> eq_elim_kw
    | NAME           -> var
\end{lstlisting}

\clearpage

\section{Mutation Taxonomy}\label{app:mutations}

\begin{table}[ht]
  \caption{{\bf Full mutation taxonomy used in the current $| \mathcal{M} | = 13$ implementation.}}
  \label{tab:mutations_list}
  \centering
  \small
  \begin{tabularx}{\textwidth}{llX}
    \toprule
    \textbf{ID} & \textbf{Class} & \textbf{Mutation name and description} \\
    \midrule
    \texttt{ID\_01} & Type Equality       & \textbf{Equality of Applications Argument} \newline Checker strictly enforces equality on both the function and the argument when comparing applications. \\
    \texttt{ID\_02} & Binding Discipline  & \textbf{Fail to Increment Under Lambda} \newline A substitution intended for an outer binder can incorrectly replace the inner bound variable. \\
    \texttt{ID\_03} & Type Equality       & \textbf{Equality of Bound Indices} \newline $\mathtt{Bound}(0)$ and $\mathtt{Bound}(1)$ are considered definitionally equal; dependent types over adjacent binders collapse. \\
    \texttt{ID\_04} & Binding Discipline  & \textbf{Index Shift} \newline $\mathtt{Bound}(i)$ lookup uses $\mathtt{env}[\min(i{+}1,\,\mathrm{len}(\mathtt{env}){-}1)]$ instead of $\mathtt{env}[i]$. Causes de Bruijn variable resolution to silently drift by one position. \\
    \texttt{ID\_05} & Semantic Environment& \textbf{Globals to Zero} \newline Unknown globals normalize to $\mathtt{Zero}$ rather than staying neutral; type checking proceeds, semantics are junk. \\
    \texttt{ID\_06} & Semantic Environment& \textbf{Lambda Environment Reversion} \newline The binder is placed at the wrong end of the environment. With the existing $\mathtt{Bound}(i) \rightarrow \mathtt{env}[i]$ convention, lambda bodies evaluate against a scrambled scope discipline. \\
    \texttt{ID\_07} & Type Equality       & \textbf{Equality of Inference Terms} \newline Equality of inference terms collapses; all types are judged equal, but inference still succeeds on well-typed terms. \\
    \texttt{ID\_08} & Semantic Environment& \textbf{Lambda Environment Padding} \newline Closure environment is padded with a neutral element, shifting all argument positions by one. \\
    \texttt{ID\_09} & Computational Rule  & \textbf{Double Application} \newline Function is applied a second time to the result of itself. \\
    \texttt{ID\_10} & Type Equality       & \textbf{Equality of Free Terms} \newline Any two neutral heads become definitionally equal after quotation; unrelated dependent types collapse. \\
    \texttt{ID\_11} & Semantic Environment& \textbf{Globals to Identity} \newline Unknown globals normalize to the identity function rather than staying neutral. \\
    \texttt{ID\_12} & Universe Leveling   & \textbf{Universe Collapse} \newline $\mathtt{Star}(n)$ returns $\mathtt{VStar}(n)$ instead of $\mathtt{VStar}(n+1)$; the universe hierarchy collapses ($\mathtt{Type}_n : \mathtt{Type}_n$). \\
    \texttt{ID\_13} & Computational Rule  & \textbf{Argument Dropping} \newline Function argument is dropped; $\mathtt{VStar}(0)$ ($\mathtt{Type}$) is used instead. \\
    \bottomrule
  \end{tabularx}
\end{table}

\begin{figure}[ht]
    \centering
    \begin{subfigure}{0.85\textwidth}
      \includegraphics[width=\textwidth]{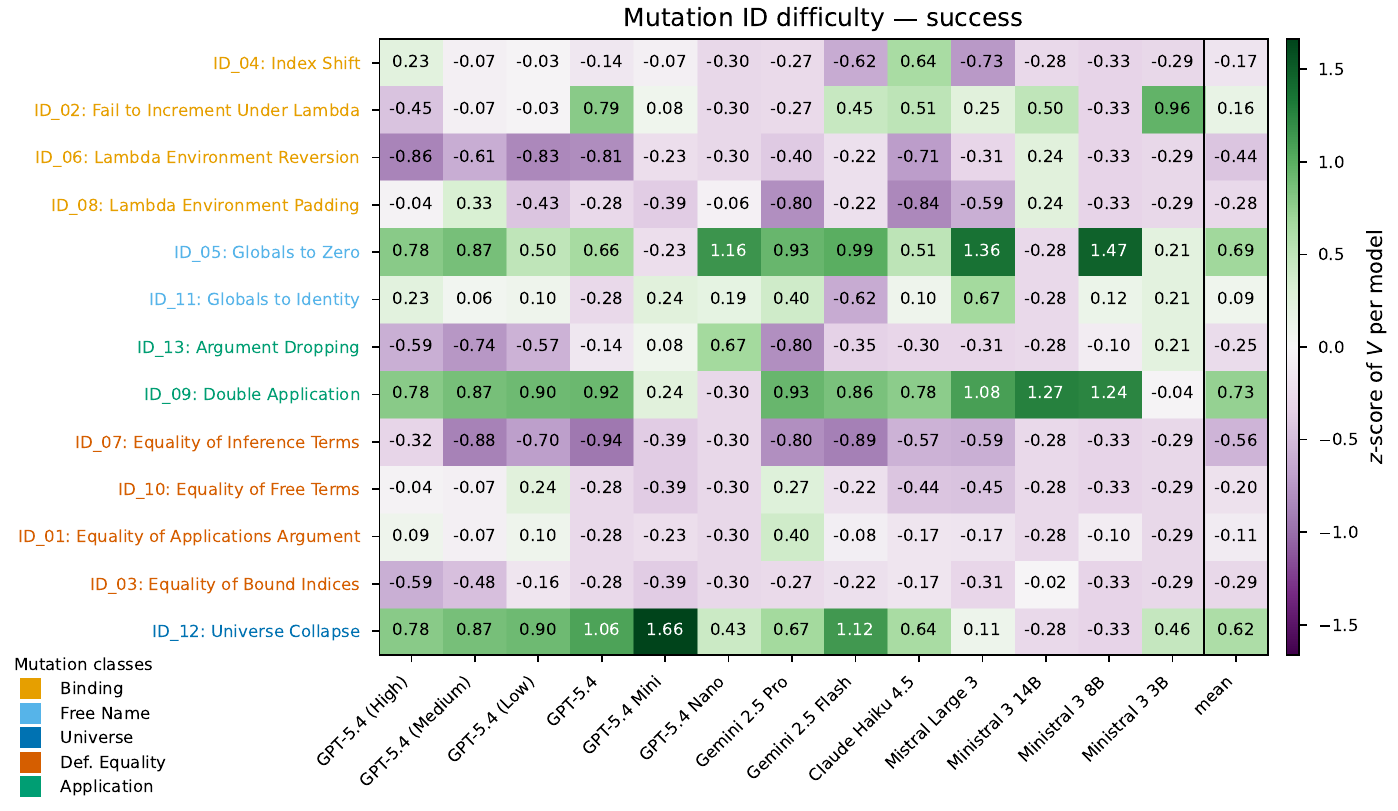}
      \caption{Per-mutation accuracy $V_K$.}
      \label{fig:mutation_difficulty_success}
    \end{subfigure}
    \begin{subfigure}{0.85\textwidth}
      \includegraphics[width=\textwidth]{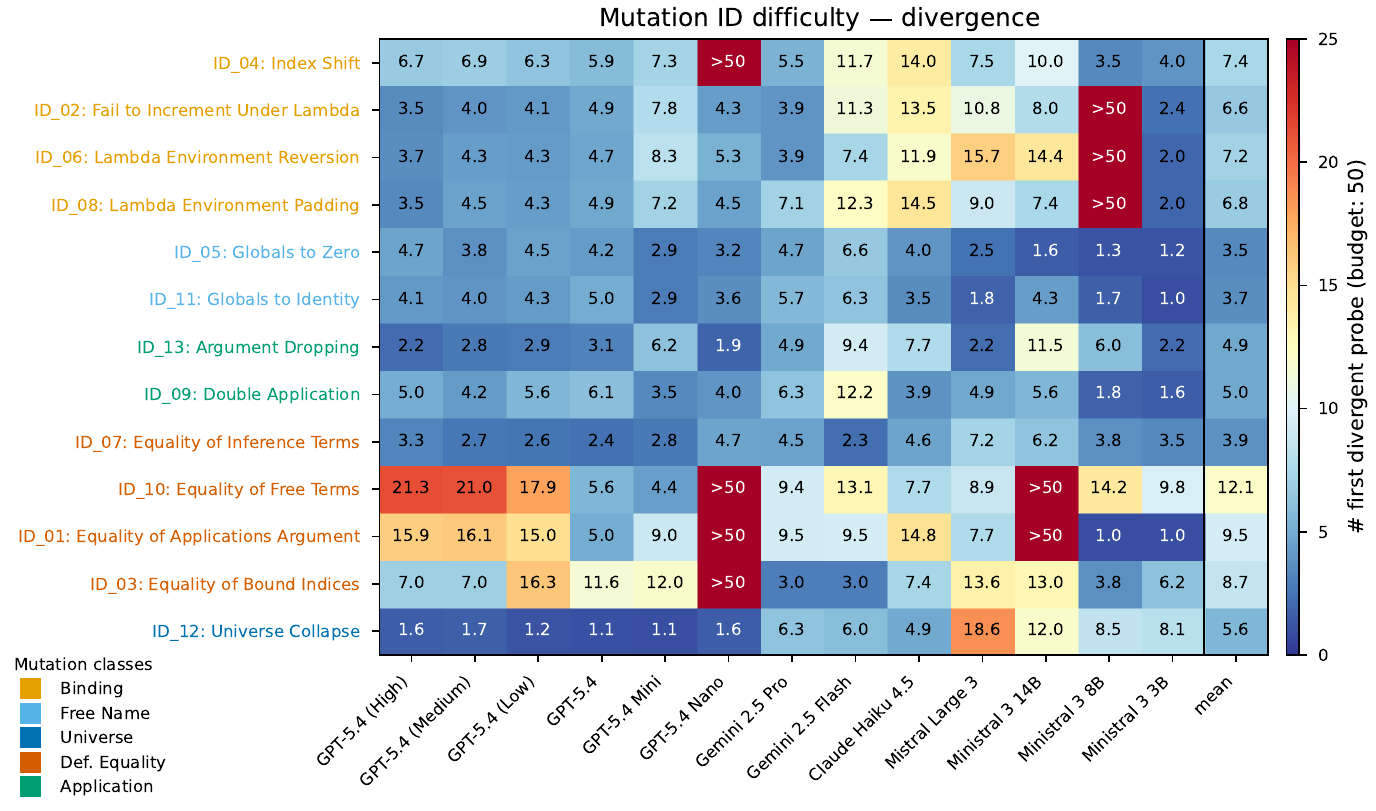}
      \caption{Per-mutation first divergent probe.}
      \label{fig:mutation_difficulty_divergence}
    \end{subfigure}
    \caption{{\bf Difficulty of $n=13$ mutations at order $k=1$ based on \enquote{Ladder} experiment.}
    {Values are averaged across prompt variant conditions.
      Classes that are consistently hard or easy to determine are shown by striking rows.
    }}
    \label{fig:mutation_difficulty}
\end{figure}

Mutation IDs rank differently in difficulty. There are mutations and classes that are inherently hard or easy regardless of model variants as observed in \autoref{fig:mutation_difficulty}.

\subsection{Class Label Mapping}\label{app:classmap}
Per-mutation experimental tables (\autoref{fig:mutation_difficulty}) use a coarser secondary grouping --- \texttt{binding}, \texttt{free\_name}, \texttt{application}, \texttt{def\_equality}, \texttt{universe} --- that was generated at experiment time.
The five-class taxonomy above (Binding Discipline, Semantic Environment, Computational Rule, Type Equality, Universe Leveling) is the canonical one used by the protocol; the coarser labels are an analysis-time grouping retained for compatibility with the released figures (\autoref{tab:mutation_classes} gives the explicit mapping).

\begin{table}[ht]
  \caption{{\bf Mapping between canonical class names and labels in figures and tables.}}
  \label{tab:mutation_classes}
  \centering
  \small
  \begin{tabularx}{\textwidth}{llX}
    \toprule
    Canonical class       & Analysis-time label     & Description \\
    \midrule
    Binding Discipline    & \texttt{binding}        & Error in variable resolution such as de Bruijn misalignment and binder shifting \\
    Semantic Environment  & \texttt{free\_name}     & Altered substitution or closure environments \\
    Computational Rule    & \texttt{application}    & Corruptions of $\beta$-reduction \\
    Type Equality         & \texttt{def\_equality}  & Changes to type-level equality or convertability \\
    Universe Leveling     & \texttt{universe}       & Violations of cumulative universe stratification \\
    \bottomrule
  \end{tabularx}
\end{table}

\subsection{Simple Reflex Agent}\label{sec:rsa}
The \texttt{Simple Reflex Agent (RSA)} is designed to diagnose all $n=13$ mutations defined above (\autoref{tab:mutations_list}) assuming singleton mutations $k=1$.
This is achieved by constructing a probe set $\Pi_\texttt{RSA} \subset \mathcal{T}$ such that each pair of distinct mutations exhibits distinguishable behaviour for at least one probe in the set, while the corresponding mutation is also distinguishable from the reference
\begin{align}
    \forall \mu,\mu' \in \mathcal{M}, \mu \neq \mu' \Rightarrow \,\exists \tau \in \Pi_\texttt{RSA} \,:\, K_{\{\mu\}}(\tau) \neq K_{\{\mu'\}}(\tau) \land K_{\{\mu\}}(\tau) \neq K_\text{ref}(\tau)
\end{align}
By probing both kernels with $\Pi_\texttt{RSA}$, the agent can uniquely identify the present mutation by comparing kernels output to the expected observation for $K_S$ and $K_\text{ref}$.\footnote{The list of constructed probes $\Pi_\texttt{RSA}$ and expected output of $K_\text{ref}$ and $K_S$ can be provided on request.} 

\subsection{Non-degeneracy Filter $k \ge 2$}\label{app:nondegen}
The candidate family for $k = 2$ is defined by
\begin{align}
    \tilde{\mathcal{F}}_2 =  \binom{\mathcal{M}}{2} \, ,
\end{align}
which are all possible 2-element combinations of mutations in $\mathcal{M}$.
However, not all of them are admissible in the sense that they can be clearly identified.
We selected the family of valid combinations $\mathcal{F}_2$ by a deterministic three-condition filter against a curated probe library based.

\paragraph{Predicate.}
Let $\Pi \subset \mathcal{T}$ denote the probe library: a fixed, ontology-balanced set of \textsc{LambdaPy} source terms that exercises every mutation class.
For a $k=2$ candidate pair $S = \{\mu_i, \mu_j\} \in \tilde{\mathcal{F}}_2$, the divergence vector $\operatorname{div}(S)$ of $K_S$ against $K_{\mathrm{ref}}$ on $\Pi$ is defined as
\begin{align}
    O_S (\tau) = (K_\text{ref}(\tau), K_S (\tau),\tau) \, , \\
    \mathrm{div}(S) = \left\lbrace O_S (\tau) : \tau \in \Pi \right\rbrace \, .
\end{align}
The pair is \emph{admissible} if
\begin{enumerate}
  \item \textbf{Non-collapse.} $\mathrm{div}(S) \neq \mathrm{div}(\emptyset)$. 
  $K_S$ has to be distinguishable from $K_\text{ref}$.
  Additionally, uniformly rejecting or uniformly accepting pairs are excluded as \enquote{total logical collapse}: $\exists\,\tau_m,\tau_n \in \Pi \text{ such that } \tau_m \neq \tau_n \land K_S(\tau_m) \neq K_S(\tau_n)$.
  \item \textbf{Non-subsumption.} $\mathrm{div}(S) \neq \mathrm{div}(\{\mu\}) \; \forall\, \mu \in \mathcal{M}$ in particular for $\mu_i,\,\mu_j$.
  Pairs whose composite divergence is indistinguishable from either singleton are excluded as \enquote{symptomatic subsumption} according to Occam's razor.
  \item \textbf{Identifiability.} $\mathrm{div}(S) \neq \mathrm{div}(S') \; \forall\,S' \in \tilde{\mathcal{F}}_2\setminus\{S\}$. Two pairs whose divergence vector is indistinguishable are excluded.
\end{enumerate}

\paragraph{Algorithm.}
For our implementation, we applied the corresponding mutation flag overrides (Section~\ref{sec:implementation}) to each of the $\binom{13}{2} = 78$ candidate unordered pairs and ran the $\Pi_\texttt{RSA}$ probe set on both $K_{\mathrm{ref}}$ and the merged kernel; each probe under a fresh state which contained only the core in \autoref{app:grammar}.
Pairs that fail condition (1) or (2) are rejected.
Those failing condition (3) are tested further by extending $\Pi_\texttt{RSA}$ with probes designed specifically to distinguish the pairs.
If this is not successful, both pairs are rejected.
The remainder is $\mathcal{F}_2$.

\paragraph{Extending to $k > 2$.}
The algorithm has to be applied iteratively for increasing $k$, where $\tilde{\mathcal{F}}_k = \binom{\mathcal{M}}{k}$.
Only condition (2) needs to be generalized for $S \in \tilde{\mathcal{F}}_k$ to $\operatorname{div}(S) \notin
\left\{\operatorname{div}(S') : S' \in \mathcal{F}_{k'},\; k' < k\right\}$.
The divergence of $S$ must be new, i.\,e. it must not coincide with the divergence of any set with lower cardinality, in particular to subsets of $S$.

\subsection{Interacting Mutation Sets}
We characterized mutation sets $S$ with cardinality $k \ge 2$ as \textit{additive} (highlighted by $k^+$) if $\forall\,\tau \in \Pi_\texttt{RSA},\ \exists\, S_\text{sub} \subsetneq S \text{ such that } \left( S_\text{sub} \text{ is additive } \lor \; |S_\text{sub}| = 1 \right) \; \land\; O_S(\tau) = O_{S_\text{sub}}(\tau)$.
This basically means, that the behaviour of a corrupted kernel $K_S$ can be fully explained by singleton mutations in $S$ and there are no unexpected results.
In contrast, we characterize mutation sets as \textit{interacting} (highlighted by $k^\otimes$) if they are not additive.
Note that there are different types of interacting sets based on a fixed probe set:
They can be partially additive (one mutation is still seen, the other not), deceptive (a singleton mutation is seen that is not active), or causing new behaviour (results can not be assigned to any known mutation).

\begin{figure}[ht]
  \centering
  \includegraphics[width=\textwidth]{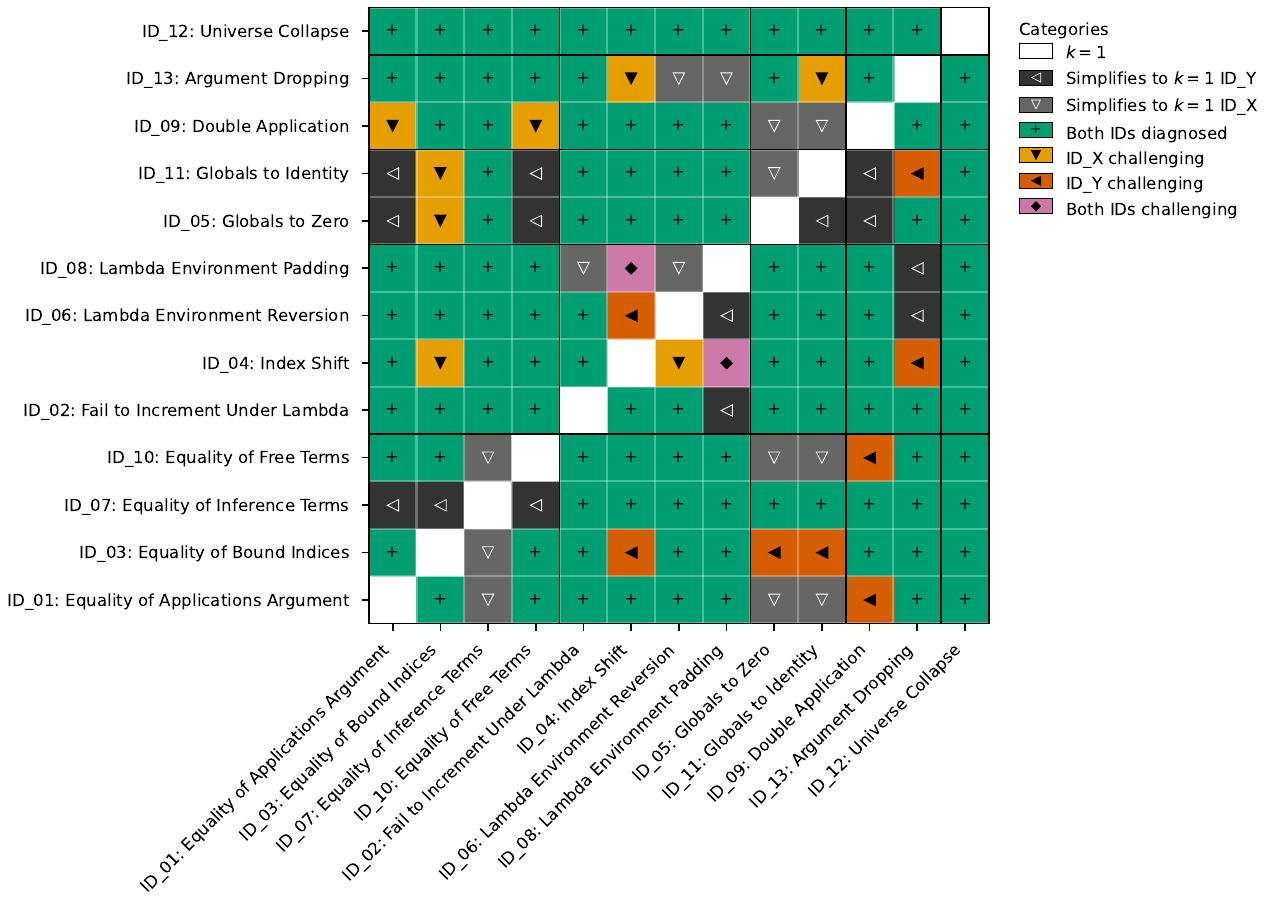}
  \caption{
    {\bf Interaction scheme of mutations.}
    Subsumed mutations (grey) are determined by i)~indication of implementation ii)~indistinguishable behaviour from $k=1$ mutations when facing $\Pi_\texttt{SRA}$.
    All combinations that are coloured are valid. Blueish green~$(+)$ is the easiest type of combination: Both mutations are additive and do not interact. There \texttt{SRA} produces $V=1$.
    Two interacting types~$(\otimes)$ are observed in our setting: i)~Partial observability (orange, red). \texttt{SRA} identifies only one mutation. Combinations kernel behaviour is still distinguishable from all others. ii)~Emergence (purple, diamond) where kernel shows a fully interacting behaviour.
  }
  \label{fig:mutation_interactions}
\end{figure}

\paragraph{Cardinalities.}
We have analyzed our \num{78} candidate pairs, see \autoref{fig:mutation_interactions}, and observed three types of mutation sets in our experimental setup, see \autoref{tab:mutation_interactions}.
\begin{table}[ht]
  \centering
  \caption{\bf Types of mutation pairs in \autoref{fig:mutation_interactions}.}
  \label{tab:mutation_interactions}
  \small
  \begin{tabular}{llr}
    \toprule
    Type & Colours in \autoref{fig:mutation_interactions} & Count \\
    \midrule
    Admissible, additive (used at $k=2^+$)  & blueish green & 55 \\
    Admissible, interacting (used at $k=2^\otimes$) & orange, red, purple  & 9 \\
    Total collapse (uniform accept/reject) & & 0 \\
    Subsumed by one of $\{\mu_i, \mu_j\}$ & light grey, dark grey & 14 \\
    \midrule
    Total                                  &  & 78 \\
    \bottomrule
  \end{tabular}
\end{table}

\subsection{Compound-Mutation Sets}
For our two test settings, Ladder and Benchmark, two different mutation families have been used.
The utilized families are listed and characterized in \autoref{tab:mutation_sets}.
\begin{table}[ht]
  \caption{\bf Curated mutation sets used in the compound-condition sessions.}
  \label{tab:mutation_sets}
  \centering
  \small
  \begin{tabular}{llccl}
    \toprule
    Order   & Mutation Set     & Ladder     &  Benchmark & Classes \\
    \midrule
    $k = 0$ & $\emptyset$        &            & \checkmark & Null baseline \\
    \midrule
    $k = 1$
    & \{\texttt{ID\_01\}}      & \checkmark &  & \texttt{def\_equality} \\
    & \{\texttt{ID\_02\}} & \checkmark & \checkmark & \texttt{binding} \\
    & \{\texttt{ID\_03\}} & \checkmark &  & \texttt{def\_equality} \\
    & \{\texttt{ID\_04\}} & \checkmark &  & \texttt{binding} \\
    & \{\texttt{ID\_05\}} & \checkmark &  & \texttt{free\_name} \\
    & \{\texttt{ID\_06\}} & \checkmark & \checkmark & \texttt{binding} \\
    & \{\texttt{ID\_07\}} & \checkmark & \checkmark & \texttt{def\_equality} \\
    & \{\texttt{ID\_08\}} & \checkmark & \checkmark & \texttt{binding} \\
    & \{\texttt{ID\_09\}} & \checkmark &  & \texttt{application} \\
    & \{\texttt{ID\_10\}} & \checkmark &  & \texttt{def\_equality} \\
    & \{\texttt{ID\_11\}} & \checkmark & \checkmark & \texttt{free\_name} \\
    & \{\texttt{ID\_12\}} & \checkmark &  & \texttt{universe} \\
    & \{\texttt{ID\_13\}} & \checkmark & \checkmark & \texttt{application} \\
    \midrule
    $k = 2^+$
    & \{\texttt{ID\_01,ID\_03\}} & \checkmark &  & \texttt{def\_equality} \\
    & \{\texttt{ID\_02,ID\_04\}} & \checkmark &  & \texttt{binding} \\
    & \{\texttt{ID\_02,ID\_06\}} & \checkmark &  & \texttt{binding} \\
    & \{\texttt{ID\_02,ID\_12\}} &  & \checkmark & \texttt{universe} $\times$ \texttt{binding} \\
    & \{\texttt{ID\_03,ID\_10\}} & \checkmark & \checkmark & \texttt{def\_equality} \\
    & \{\texttt{ID\_04,ID\_09\}} & \checkmark &  & \texttt{application} $\times$ \texttt{binding} \\
    & \{\texttt{ID\_06,ID\_09\}} & \checkmark &  & \texttt{application} $\times$ \texttt{binding} \\
    & \{\texttt{ID\_06,ID\_10\}} & \checkmark &  & \texttt{def\_equality} $\times$ \texttt{binding} \\
    & \{\texttt{ID\_07,ID\_13\}} &  & \checkmark & \texttt{def\_equality} $\times$ \texttt{application} \\
    \midrule
    $k = 2^\otimes$
    & \{\texttt{ID\_04,ID\_06\}} & \checkmark & \checkmark & \texttt{binding} \\
    & \{\texttt{ID\_04,ID\_08\}} & \checkmark &  & \texttt{binding} \\
    & \{\texttt{ID\_09,ID\_10\}} & \checkmark &  & \texttt{def\_equality} $\times$ \texttt{application} \\
    \midrule
    $k = 3^+$
    & \{\texttt{ID\_01,ID\_03,ID\_10\}} &  & \checkmark & \texttt{def\_equality} \\
    & \{\texttt{ID\_02,ID\_03,ID\_06\}} &  & \checkmark & \texttt{def\_equality} $\times$ \texttt{binding} \\
     & \{\texttt{ID\_07,ID\_09,ID\_12\}} &  & \checkmark & \texttt{def\_equality} $\times$ \texttt{application} $\times$ \texttt{universe} \\
    \bottomrule
  \end{tabular}
\end{table}

\clearpage

\section{Detailed Model-by-condition Tables}\label{app:detail_tables}

The Section~\ref{sec:experiments} main text reports only the headline inline, e.\,g. \autoref{tab:det_attr_gap}.
All other per-model and per-condition tables are collected here.
The wide tables are set on their own landscape pages.

\paragraph{\enquote{Ladder} experiment.} \autoref{tab:multi_mutation_effect} marginalises over scaffold to expose the single\,vs.\,double dissociation as well as the interacting difficulty.
Similarly, \autoref{tab:hint_effect} summarizes the effect of hint levels in the system prompt per model.
\autoref{tab:reporting} expands the over/under-reporting breakdown at $k=2$.

\begin{table}[ht]
  \centering
  \caption{{\bf Per-model comparison of single\,vs.\,double\,vs.\,interacting mutations.}
  {Marginal accuracy by mutation count $k$ on the \enquote{Ladder} experiment: exact correctness $V$, kernel detection $V_K$, exact attribution $V_A$, Jaccard partial credit $V_A^{J}$, conditional attribution correctness $V_A | V_K = 1$, and ratio of used probes $P / P_\text{max}$.
  Numbers are mean values over prompt levels and corresponding mutations sets.
  For $k \ge 2$, superscripts indicate whether the mutations are additive~$(+)$ or interacting~$(\otimes)$.
  }}
\begin{tabular}{llrrrrrrr}
\toprule
Model & $k$ & $n$ & $V$ & $V_K$ & $V_A$ & $V_A^{J}$ & $V_A \mid V_K$ & $P / P_\text{max}$  \\
\midrule
\texttt{ministral-3b-2512} & $1$ & 195 & 0.077 & 0.369 & 0.190 & 0.190 & 0.208 & 0.616 \\
 & $2^+$ & 105 & 0.000 & 0.295 & 0.000 & 0.094 & 0.000 & 0.727 \\
 & $2^\otimes$ & 45 & 0.000 & 0.267 & 0.000 & 0.015 & 0.000 & 0.775 \\
\addlinespace[0.33em] 
\texttt{ministral-8b-2512} & $1$ & 195 & 0.097 & 0.210 & 0.200 & 0.200 & 0.463 & 0.966 \\
 & $2^+$ & 105 & 0.010 & 0.238 & 0.067 & 0.160 & 0.040 & 0.988 \\
 & $2^\otimes$ & 45 & 0.000 & 0.200 & 0.000 & 0.074 & 0.000 & 0.980 \\
\addlinespace[0.33em] 
\texttt{ministral-14b-2512} & $1$ & 195 & 0.072 & 0.272 & 0.082 & 0.085 & 0.264 & 0.999 \\
 & $2^+$ & 105 & 0.010 & 0.352 & 0.019 & 0.122 & 0.027 & 1.000 \\
 & $2^\otimes$ & 45 & 0.000 & 0.356 & 0.000 & 0.096 & 0.000 & 1.000 \\
\addlinespace[0.33em] 
\texttt{mistral-large-2512} & $1$ & 195 & 0.349 & 0.600 & 0.451 & 0.451 & 0.581 & 0.573 \\
 & $2^+$ & 105 & 0.133 & 0.638 & 0.171 & 0.363 & 0.209 & 0.619 \\
 & $2^\otimes$ & 45 & 0.044 & 0.867 & 0.044 & 0.274 & 0.051 & 0.550 \\
\addlinespace[0.33em] 
\texttt{claude-haiku-4-5} & $1$ & 195 & 0.415 & 0.774 & 0.451 & 0.456 & 0.536 & 0.763 \\
 & $2^+$ & 105 & 0.181 & 0.743 & 0.229 & 0.386 & 0.244 & 0.813 \\
 & $2^\otimes$ & 45 & 0.133 & 0.911 & 0.133 & 0.404 & 0.146 & 0.852 \\
\addlinespace[0.33em] 
\texttt{gemini-2.5-flash} & $1$ & 195 & 0.441 & 0.877 & 0.446 & 0.459 & 0.503 & 0.378 \\
 & $2^+$ & 105 & 0.295 & 0.905 & 0.295 & 0.471 & 0.326 & 0.470 \\
 & $2^\otimes$ & 45 & 0.156 & 0.911 & 0.156 & 0.393 & 0.171 & 0.427 \\
\addlinespace[0.33em] 
\texttt{gemini-2.5-pro} & $1$ & 195 & 0.467 & 0.918 & 0.467 & 0.495 & 0.508 & 0.342 \\
 & $2^+$ & 105 & 0.200 & 0.933 & 0.210 & 0.379 & 0.214 & 0.434 \\
 & $2^\otimes$ & 45 & 0.200 & 0.956 & 0.200 & 0.430 & 0.209 & 0.419 \\
\addlinespace[0.33em] 
\texttt{gpt-5.4-nano} & $1$ & 195 & 0.082 & 0.174 & 0.133 & 0.133 & 0.471 & 0.153 \\
 & $2^+$ & 105 & 0.000 & 0.124 & 0.000 & 0.083 & 0.000 & 0.170 \\
 & $2^\otimes$ & 45 & 0.000 & 0.200 & 0.000 & 0.000 & 0.000 & 0.152 \\
\addlinespace[0.33em] 
\texttt{gpt-5.4-mini} & $1$ & 195 & 0.231 & 0.569 & 0.318 & 0.326 & 0.405 & 0.304 \\
 & $2^+$ & 105 & 0.076 & 0.581 & 0.152 & 0.298 & 0.131 & 0.328 \\
 & $2^\otimes$ & 45 & 0.022 & 0.622 & 0.022 & 0.211 & 0.036 & 0.232 \\
\addlinespace[0.33em] 
\texttt{gpt-5.4} & $1$ & 195 & 0.472 & 0.851 & 0.472 & 0.487 & 0.554 & 0.310 \\
 & $2^+$ & 105 & 0.267 & 0.848 & 0.267 & 0.386 & 0.315 & 0.348 \\
 & $2^\otimes$ & 45 & 0.133 & 0.956 & 0.133 & 0.396 & 0.140 & 0.325 \\
\addlinespace[0.33em] 
\texttt{gpt-5.4\_low} & $1$ & 195 & 0.549 & 0.933 & 0.549 & 0.574 & 0.588 & 0.265 \\
 & $2^+$ & 105 & 0.267 & 0.914 & 0.267 & 0.421 & 0.292 & 0.347 \\
 & $2^\otimes$ & 45 & 0.178 & 1.000 & 0.178 & 0.478 & 0.178 & 0.295 \\
\addlinespace[0.33em] 
\texttt{gpt-5.4\_medium} & $1$ & 195 & 0.569 & 0.908 & 0.569 & 0.585 & 0.627 & 0.311 \\
 & $2^+$ & 105 & 0.362 & 0.905 & 0.362 & 0.524 & 0.400 & 0.449 \\
 & $2^\otimes$ & 45 & 0.289 & 1.000 & 0.289 & 0.507 & 0.289 & 0.404 \\
\addlinespace[0.33em] 
\texttt{gpt-5.4\_high} & $1$ & 195 & 0.621 & 0.913 & 0.621 & 0.628 & 0.680 & 0.326 \\
 & $2^+$ & 105 & 0.276 & 0.848 & 0.276 & 0.400 & 0.326 & 0.484 \\
 & $2^\otimes$ & 45 & 0.178 & 0.889 & 0.178 & 0.419 & 0.200 & 0.461 \\
\bottomrule
\end{tabular}

  \label{tab:multi_mutation_effect}
\end{table}

\begin{table}[ht]
  \centering
  \caption{{\bf Per-model comparison of scaffolding level.}
  {Marginal accuracy by prompt level on the \enquote{Ladder} experiment: exact correctness $V$, kernel detection $V_K$, exact attribution $V_A$, Jaccard partial credit $V_A^{J}$, conditional attribution correctness $V_A | V_K = 1$, and ratio of used probes $P / P_\text{max}$.
  Numbers are mean values over mutations sets.
  }}
\begin{tabular}{llrrrrrrr}
\toprule
Model & Prompt & $n$ & $V$ & $V_K$ & $V_A$ & $V_A^{J}$ & $V_A \mid V_K$ & $P / P_\text{max}$  \\
\midrule
\texttt{ministral-3b-2512} & Level 0 & 115 & 0.113 & 0.470 & 0.287 & 0.378 & 0.241 & 0.206 \\
 & Level 1 & 115 & 0.009 & 0.217 & 0.009 & 0.009 & 0.040 & 0.874 \\
 & Level 2 & 115 & 0.009 & 0.313 & 0.026 & 0.026 & 0.028 & 0.932 \\
\addlinespace[0.33em] 
\texttt{ministral-8b-2512} & Level 0 & 115 & 0.078 & 0.174 & 0.287 & 0.336 & 0.450 & 0.936 \\
 & Level 1 & 115 & 0.035 & 0.296 & 0.043 & 0.078 & 0.118 & 0.991 \\
 & Level 2 & 115 & 0.061 & 0.183 & 0.070 & 0.100 & 0.333 & 0.996 \\
\addlinespace[0.33em] 
\texttt{ministral-14b-2512} & Level 0 & 115 & 0.061 & 0.339 & 0.087 & 0.130 & 0.179 & 1.000 \\
 & Level 1 & 115 & 0.035 & 0.191 & 0.035 & 0.058 & 0.182 & 0.998 \\
 & Level 2 & 115 & 0.035 & 0.391 & 0.035 & 0.104 & 0.089 & 1.000 \\
\addlinespace[0.33em] 
\texttt{mistral-large-2512} & Level 0 & 115 & 0.409 & 0.791 & 0.539 & 0.620 & 0.516 & 0.331 \\
 & Level 1 & 115 & 0.130 & 0.496 & 0.157 & 0.223 & 0.263 & 0.749 \\
 & Level 2 & 115 & 0.191 & 0.652 & 0.243 & 0.361 & 0.293 & 0.672 \\
\addlinespace[0.33em] 
\texttt{claude-haiku-4-5} & Level 0 & 115 & 0.548 & 0.939 & 0.609 & 0.678 & 0.583 & 0.679 \\
 & Level 1 & 115 & 0.252 & 0.722 & 0.261 & 0.339 & 0.349 & 0.832 \\
 & Level 2 & 115 & 0.122 & 0.687 & 0.157 & 0.267 & 0.177 & 0.859 \\
\addlinespace[0.33em] 
\texttt{gemini-2.5-flash} & Level 0 & 115 & 0.539 & 0.957 & 0.539 & 0.583 & 0.564 & 0.172 \\
 & Level 1 & 115 & 0.278 & 0.896 & 0.287 & 0.359 & 0.311 & 0.475 \\
 & Level 2 & 115 & 0.261 & 0.817 & 0.261 & 0.420 & 0.319 & 0.590 \\
\addlinespace[0.33em] 
\texttt{gemini-2.5-pro} & Level 0 & 115 & 0.557 & 0.965 & 0.557 & 0.591 & 0.577 & 0.161 \\
 & Level 1 & 115 & 0.339 & 0.922 & 0.348 & 0.429 & 0.368 & 0.458 \\
 & Level 2 & 115 & 0.157 & 0.896 & 0.157 & 0.333 & 0.175 & 0.520 \\
\addlinespace[0.33em] 
\texttt{gpt-5.4-nano} & Level 0 & 115 & 0.113 & 0.209 & 0.191 & 0.214 & 0.542 & 0.111 \\
 & Level 1 & 115 & 0.009 & 0.139 & 0.009 & 0.061 & 0.062 & 0.151 \\
 & Level 2 & 115 & 0.017 & 0.139 & 0.026 & 0.026 & 0.125 & 0.212 \\
\addlinespace[0.33em] 
\texttt{gpt-5.4-mini} & Level 0 & 115 & 0.270 & 0.687 & 0.383 & 0.470 & 0.392 & 0.141 \\
 & Level 1 & 115 & 0.113 & 0.548 & 0.183 & 0.246 & 0.206 & 0.353 \\
 & Level 2 & 115 & 0.087 & 0.504 & 0.122 & 0.191 & 0.172 & 0.411 \\
\addlinespace[0.33em] 
\texttt{gpt-5.4} & Level 0 & 115 & 0.617 & 1.000 & 0.617 & 0.670 & 0.617 & 0.176 \\
 & Level 1 & 115 & 0.261 & 0.791 & 0.261 & 0.328 & 0.330 & 0.371 \\
 & Level 2 & 115 & 0.217 & 0.800 & 0.217 & 0.336 & 0.272 & 0.424 \\
\addlinespace[0.33em] 
\texttt{gpt-5.4\_low} & Level 0 & 115 & 0.548 & 1.000 & 0.548 & 0.629 & 0.548 & 0.174 \\
 & Level 1 & 115 & 0.443 & 0.904 & 0.443 & 0.519 & 0.490 & 0.346 \\
 & Level 2 & 115 & 0.252 & 0.904 & 0.252 & 0.397 & 0.279 & 0.362 \\
\addlinespace[0.33em] 
\texttt{gpt-5.4\_medium} & Level 0 & 115 & 0.591 & 1.000 & 0.591 & 0.667 & 0.591 & 0.211 \\
 & Level 1 & 115 & 0.443 & 0.896 & 0.443 & 0.504 & 0.495 & 0.423 \\
 & Level 2 & 115 & 0.374 & 0.861 & 0.374 & 0.497 & 0.434 & 0.462 \\
\addlinespace[0.33em] 
\texttt{gpt-5.4\_high} & Level 0 & 115 & 0.635 & 0.974 & 0.635 & 0.693 & 0.652 & 0.231 \\
 & Level 1 & 115 & 0.400 & 0.817 & 0.400 & 0.446 & 0.489 & 0.446 \\
 & Level 2 & 115 & 0.339 & 0.878 & 0.339 & 0.455 & 0.386 & 0.497 \\
\bottomrule
\end{tabular}

  \label{tab:hint_effect}
\end{table}

\begin{table}[ht]
  \centering
  \caption{{\bf Fault characterization for $k=2$ double mutation in the \enquote{Ladder} experiment.}
  {Per-model and $k$-hint level prompts, attribution modes are analyzed for $n$ sessions with $V_K = 1$.
  Each entry shows the proportion of responses that were classified as correct, incomplete (partially correct but missing mutations), mixed (partially correct but containing incorrect mutations), or incorrect.
  Proportions sum to one for each model and prompt.
  $<P_\text{max}$ gives the proportion of sessions with probe budget NOT exhausted although incorrect mutation attribution $V_K = 1 \land V_A = 0$.
  }}
\begin{tabular}{llrrrrrr}
\toprule
Model & Prompt & $n$ & Correct & Incomplete & Mixed & Incorrect & $<P_\text{max}$  \\
\midrule
\texttt{ministral-3b-2512} & Level 0-1 & 26 & 0.000 & 0.192 & 0.192 & 0.615 & 0.731 \\
 & Level 2 & 17 & 0.000 & 0.000 & 0.000 & 1.000 & 0.118 \\
\addlinespace[0.33em] 
\texttt{ministral-8b-2512} & Level 0-1 & 25 & 0.040 & 0.000 & 0.680 & 0.280 & 0.208 \\
 & Level 2 & 9 & 0.000 & 0.778 & 0.000 & 0.222 & 0.222 \\
\addlinespace[0.33em] 
\texttt{ministral-14b-2512} & Level 0-1 & 29 & 0.034 & 0.138 & 0.483 & 0.345 & 0.000 \\
 & Level 2 & 24 & 0.000 & 0.625 & 0.000 & 0.375 & 0.000 \\
\addlinespace[0.33em] 
\texttt{mistral-large-2512} & Level 0-1 & 70 & 0.229 & 0.000 & 0.643 & 0.129 & 0.926 \\
 & Level 2 & 36 & 0.000 & 0.750 & 0.000 & 0.250 & 0.917 \\
\addlinespace[0.33em] 
\texttt{claude-haiku-4-5} & Level 0-1 & 82 & 0.305 & 0.000 & 0.573 & 0.122 & 0.965 \\
 & Level 2 & 37 & 0.000 & 0.541 & 0.135 & 0.324 & 0.946 \\
\addlinespace[0.33em] 
\texttt{gemini-2.5-flash} & Level 0-1 & 92 & 0.359 & 0.000 & 0.435 & 0.207 & 0.898 \\
 & Level 2 & 44 & 0.114 & 0.705 & 0.023 & 0.159 & 0.923 \\
\addlinespace[0.33em] 
\texttt{gemini-2.5-pro} & Level 0-1 & 93 & 0.323 & 0.000 & 0.430 & 0.247 & 0.889 \\
 & Level 2 & 48 & 0.000 & 0.479 & 0.208 & 0.312 & 0.979 \\
\addlinespace[0.33em] 
\texttt{gpt-5.4-nano} & Level 0-1 & 12 & 0.000 & 0.000 & 0.417 & 0.583 & 1.000 \\
 & Level 2 & 10 & 0.000 & 0.000 & 0.000 & 1.000 & 1.000 \\
\addlinespace[0.33em] 
\texttt{gpt-5.4-mini} & Level 0-1 & 65 & 0.138 & 0.000 & 0.569 & 0.292 & 1.000 \\
 & Level 2 & 24 & 0.000 & 0.292 & 0.083 & 0.625 & 0.958 \\
\addlinespace[0.33em] 
\texttt{gpt-5.4} & Level 0-1 & 91 & 0.319 & 0.000 & 0.451 & 0.231 & 1.000 \\
 & Level 2 & 41 & 0.122 & 0.390 & 0.195 & 0.293 & 1.000 \\
\addlinespace[0.33em] 
\texttt{gpt-5.4\_low} & Level 0-1 & 96 & 0.323 & 0.000 & 0.552 & 0.125 & 1.000 \\
 & Level 2 & 45 & 0.111 & 0.444 & 0.111 & 0.333 & 1.000 \\
\addlinespace[0.33em] 
\texttt{gpt-5.4\_medium} & Level 0-1 & 95 & 0.432 & 0.000 & 0.484 & 0.084 & 0.981 \\
 & Level 2 & 45 & 0.222 & 0.378 & 0.178 & 0.222 & 0.971 \\
\addlinespace[0.33em] 
\texttt{gpt-5.4\_high} & Level 0-1 & 85 & 0.400 & 0.000 & 0.424 & 0.176 & 0.922 \\
 & Level 2 & 44 & 0.068 & 0.386 & 0.227 & 0.318 & 0.976 \\
\midrule
Overall & Level 0--2 & 1285 & 0.216 & 0.163 & 0.370 & 0.251 & 0.865 \\
\bottomrule
\end{tabular}

  \label{tab:reporting}
\end{table}

\paragraph{\enquote{Benchmark} experiment.} \autoref{tab:null_baseline} lists per-model true-negative rates and probing perseverance on the $k=0$ null condition.
\autoref{tab:full_benckmark} gives the full per-(model, condition) summary of detailed performance statistics for all benchmark configurations with for $k\in\{1,2,3\}$.

\begin{table}[ht]
\centering
\caption{{\bf $k=0$ null-condition coverage in the \enquote{Benchmark} experiment.}
  {Per-model true-negative rate (specificity), number of probes used, and utilization of probe budget.
Point estimates only are given because each per-model cell is $n=5$, too few for a useful interval.
Models are sorted as in \autoref{fig:leaderboard}.
The bottom row aggregates across all $155$ null sessions.}}
\newcommand{\scorebar}[1]{%
\begin{tikzpicture}[baseline=-0.6ex]
  \fill[gray!20] (0,0) rectangle (2,0.18);
  \fill[blue] (0,0) rectangle ({2*#1},0.18);
\end{tikzpicture}
}
\begin{tabular}{lrrrrrc}
\toprule
Model & $n$ & TN & FP & Specificity & Probes & Budget exhausted \\
\midrule
\texttt{ministral-3b-2512} & 5 & 1 & 4 & 0.20 & 65.0 & \scorebar{1.000} \\
\texttt{magistral-small-2509} & 5 & 2 & 3 & 0.40 & 28.0 & \scorebar{0.431} \\
\texttt{ministral-8b-2512} & 5 & 5 & 0 & 1.00 & 65.0 & \scorebar{1.000} \\
\texttt{mistral-medium-2508} & 5 & 5 & 0 & 1.00 & 65.0 & \scorebar{1.000} \\
\texttt{ministral-14b-2512} & 5 & 5 & 0 & 1.00 & 65.0 & \scorebar{1.000} \\
\texttt{mistral-small-2603} & 5 & 5 & 0 & 1.00 & 65.0 & \scorebar{1.000} \\
\texttt{gemini-2.5-pro} & 5 & 4 & 1 & 0.80 & 61.0 & \scorebar{0.938} \\
\texttt{gpt-5.4-nano} & 5 & 5 & 0 & 1.00 & 6.4 & \scorebar{0.098} \\
\texttt{gpt-5.4-mini} & 5 & 5 & 0 & 1.00 & 15.2 & \scorebar{0.234} \\
\texttt{gpt-5.4\_xhigh} & 5 & 2 & 3 & 0.40 & 27.4 & \scorebar{0.422} \\
\texttt{mistral-medium-3-5} & 5 & 3 & 2 & 0.60 & 65.0 & \scorebar{1.000} \\
\texttt{gemini-2.5-flash} & 5 & 3 & 2 & 0.60 & 62.8 & \scorebar{0.966} \\
\texttt{mistral-large-2512} & 5 & 4 & 1 & 0.80 & 65.0 & \scorebar{1.000} \\
\texttt{claude-haiku-4-5} & 5 & 5 & 0 & 1.00 & 63.8 & \scorebar{0.982} \\
\texttt{gpt-5.4} & 5 & 5 & 0 & 1.00 & 26.6 & \scorebar{0.409} \\
\texttt{magistral-medium-2509} & 5 & 5 & 0 & 1.00 & 39.2 & \scorebar{0.603} \\
\texttt{gpt-5.4\_high} & 5 & 5 & 0 & 1.00 & 26.0 & \scorebar{0.400} \\
\texttt{gpt-5.5} & 5 & 5 & 0 & 1.00 & 33.6 & \scorebar{0.517} \\
\texttt{gpt-5.4\_low} & 5 & 5 & 0 & 1.00 & 20.4 & \scorebar{0.314} \\
\texttt{claude-sonnet-4-6} & 5 & 5 & 0 & 1.00 & 65.0 & \scorebar{1.000} \\
\texttt{gpt-5.5\_low} & 5 & 5 & 0 & 1.00 & 28.4 & \scorebar{0.437} \\
\texttt{gpt-5.4\_medium} & 5 & 5 & 0 & 1.00 & 22.4 & \scorebar{0.345} \\
\texttt{gpt-5.5\_medium} & 5 & 5 & 0 & 1.00 & 41.2 & \scorebar{0.634} \\
\texttt{gpt-5.5\_high} & 5 & 5 & 0 & 1.00 & 52.4 & \scorebar{0.806} \\
\texttt{gpt-5.5\_xhigh} & 5 & 5 & 0 & 1.00 & 50.0 & \scorebar{0.769} \\
\texttt{claude-sonnet-4-6\_low} & 5 & 5 & 0 & 1.00 & 63.2 & \scorebar{0.972} \\
\texttt{claude-sonnet-4-6\_high} & 5 & 5 & 0 & 1.00 & 61.0 & \scorebar{0.938} \\
\texttt{claude-opus-4-7\_high} & 5 & 5 & 0 & 1.00 & 54.0 & \scorebar{0.831} \\
\texttt{gemini-3.1-pro-preview} & 5 & 5 & 0 & 1.00 & 56.2 & \scorebar{0.865} \\
\texttt{gemini-3.5-flash} & 5 & 5 & 0 & 1.00 & 64.0 & \scorebar{0.985} \\
\texttt{simple\_reflex\_agent} & 5 & 5 & 0 & 1.00 & 14.0 & \scorebar{0.215} \\
\midrule
Overall & 155 & 139 & 16 & 0.90 & 46.4 & \scorebar{0.713} \\
\bottomrule
\end{tabular}

\label{tab:null_baseline}
\end{table}

\begin{landscape}
\small
\centering
\begin{longtable}{llrrrrrrrrrrr}
  \caption{\textbf{Summary statistics per model and experiment setting on \enquote{Benchmark}.}
  V is exact-match correctness; VAJ is the Jaccard partial-credit score on the mutation set.
  $k$ consists of $n$ mutation sets.
  Metrics are exact correctness $V$, kernel detection $V_K$, exact attribution $V_A$, Jaccard partial credit $V_A^{J}$, conditional attribution correctness $V_A | V_K = 1$, precision and recall of attribution, and probing efficiency respresented by $E_\text{ELX}$, ratio of used probes $P / P_\text{max}$, and ratio of in probe budget.
  Numbers are averaged values per model and $k$ across $n$ mutation sets.
  }
  \label{tab:full_benckmark} \\
  \toprule
  Model & $k$ & $n$ & $V$ & $V_K$ & $V_A$ & $V_A^{J}$ & $V_A \mid V_K$ & $\text{Precision}_A$ & $\text{Recall}_A$ & $E_\text{ELX}$ & $P / P_\text{max}$ & $<P_\text{max}$ \\
  \midrule
  \endfirsthead
  \caption[]{(continued)} \\
  \toprule
   Model & $k$ & $n$ & $V$ & $V_K$ & $V_A$ & $V_A^{J}$ & $V_A \mid V_K$ & $\text{Precision}_A$ & $\text{Recall}_A$ & $E_\text{ELX}$ & $P / P_\text{max}$ & $<P_\text{max}$ \\
  \midrule
  \endhead
  \midrule
  \multicolumn{13}{r}{Continued on next page} \\
  \midrule
  \endfoot
  \bottomrule
  \endlastfoot
\multirow{4}{*}{}{\texttt{claude-haiku-4-5}} & $1$ & 30 & 0.30 & 0.97 & 0.30 & 0.33 & 0.31 & 0.33 & 0.37 & 0.37 & 0.83 & 0.90 \\*
 & $2^+$ & 15 & 0.00 & 0.47 & 0.00 & 0.17 & 0.00 & 0.17 & 0.17 & 0.00 & 0.79 & 0.80 \\*
 & $3^+$ & 15 & 0.00 & 0.87 & 0.00 & 0.22 & 0.00 & 0.22 & 0.22 & 0.00 & 0.78 & 1.00 \\*
 & $2^\otimes$ & 5 & 0.20 & 1.00 & 0.20 & 0.50 & 0.20 & 0.50 & 0.50 & 0.24 & 0.88 & 1.00 \\
\addlinespace[0.33em] 
\multirow{4}{*}{}{\texttt{claude-sonnet-4-6}} & $1$ & 30 & 0.33 & 1.00 & 0.33 & 0.53 & 0.33 & 0.53 & 0.73 & 0.33 & 1.00 & 0.00 \\*
 & $2^+$ & 15 & 0.33 & 0.87 & 0.33 & 0.46 & 0.38 & 0.46 & 0.47 & 0.33 & 1.00 & 0.07 \\*
 & $3^+$ & 15 & 0.00 & 0.80 & 0.00 & 0.41 & 0.00 & 0.41 & 0.42 & 0.00 & 1.00 & 0.07 \\*
 & $2^\otimes$ & 5 & 0.40 & 1.00 & 0.40 & 0.67 & 0.40 & 0.67 & 0.70 & 0.40 & 1.00 & 0.00 \\
\addlinespace[0.33em] 
\multirow{4}{*}{}{\texttt{claude-sonnet-4-6\_low}} & $1$ & 30 & 0.57 & 1.00 & 0.57 & 0.62 & 0.57 & 0.62 & 0.67 & 0.78 & 0.59 & 0.93 \\*
 & $2^+$ & 15 & 0.33 & 0.67 & 0.33 & 0.39 & 0.50 & 0.39 & 0.40 & 0.46 & 0.77 & 0.60 \\*
 & $3^+$ & 15 & 0.20 & 0.87 & 0.20 & 0.46 & 0.23 & 0.46 & 0.47 & 0.28 & 0.67 & 0.87 \\*
 & $2^\otimes$ & 5 & 0.00 & 1.00 & 0.00 & 0.50 & 0.00 & 0.50 & 0.50 & 0.00 & 0.69 & 1.00 \\
\addlinespace[0.33em] 
\multirow{4}{*}{}{\texttt{claude-sonnet-4-6\_high}} & $1$ & 30 & 0.67 & 1.00 & 0.67 & 0.72 & 0.67 & 0.72 & 0.77 & 0.91 & 0.62 & 1.00 \\*
 & $2^+$ & 15 & 0.33 & 0.67 & 0.33 & 0.36 & 0.50 & 0.36 & 0.37 & 0.45 & 0.72 & 0.80 \\*
 & $3^+$ & 15 & 0.27 & 0.87 & 0.27 & 0.57 & 0.31 & 0.57 & 0.58 & 0.37 & 0.68 & 0.87 \\*
 & $2^\otimes$ & 5 & 0.00 & 1.00 & 0.00 & 0.50 & 0.00 & 0.50 & 0.50 & 0.00 & 0.60 & 1.00 \\
\addlinespace[0.33em] 
\multirow{4}{*}{}{\texttt{claude-opus-4-7\_high}} & $1$ & 30 & 0.83 & 1.00 & 0.83 & 0.83 & 0.83 & 0.83 & 0.83 & 1.21 & 0.42 & 1.00 \\*
 & $2^+$ & 15 & 0.27 & 0.93 & 0.27 & 0.51 & 0.29 & 0.51 & 0.57 & 0.40 & 0.45 & 1.00 \\*
 & $3^+$ & 15 & 0.13 & 1.00 & 0.13 & 0.51 & 0.13 & 0.51 & 0.53 & 0.20 & 0.35 & 1.00 \\*
 & $2^\otimes$ & 5 & 0.00 & 1.00 & 0.00 & 0.50 & 0.00 & 0.50 & 0.50 & 0.00 & 0.68 & 1.00 \\
\addlinespace[0.33em] 
\multirow{4}{*}{}{\texttt{gemini-2.5-flash}} & $1$ & 30 & 0.23 & 0.93 & 0.23 & 0.26 & 0.25 & 0.26 & 0.30 & 0.33 & 0.42 & 1.00 \\*
 & $2^+$ & 15 & 0.00 & 0.73 & 0.00 & 0.19 & 0.00 & 0.19 & 0.20 & 0.00 & 0.43 & 1.00 \\*
 & $3^+$ & 15 & 0.00 & 0.73 & 0.00 & 0.20 & 0.00 & 0.20 & 0.27 & 0.00 & 0.49 & 0.93 \\*
 & $2^\otimes$ & 5 & 0.20 & 0.80 & 0.20 & 0.30 & 0.25 & 0.30 & 0.30 & 0.29 & 0.41 & 1.00 \\
\addlinespace[0.33em] 
\multirow{4}{*}{}{\texttt{gemini-2.5-pro}} & $1$ & 30 & 0.10 & 0.97 & 0.10 & 0.19 & 0.10 & 0.19 & 0.30 & 0.15 & 0.32 & 1.00 \\*
 & $2^+$ & 15 & 0.00 & 0.87 & 0.00 & 0.22 & 0.00 & 0.22 & 0.27 & 0.00 & 0.37 & 1.00 \\*
 & $3^+$ & 15 & 0.00 & 0.93 & 0.00 & 0.39 & 0.00 & 0.39 & 0.44 & 0.00 & 0.35 & 1.00 \\*
 & $2^\otimes$ & 5 & 0.20 & 1.00 & 0.20 & 0.50 & 0.20 & 0.50 & 0.50 & 0.29 & 0.25 & 1.00 \\
\addlinespace[0.33em] 
\multirow{4}{*}{}{\texttt{gemini-3.1-pro-preview}} & $1$ & 30 & 0.80 & 1.00 & 0.80 & 0.80 & 0.80 & 0.80 & 0.80 & 1.10 & 0.58 & 1.00 \\*
 & $2^+$ & 15 & 0.33 & 1.00 & 0.33 & 0.61 & 0.33 & 0.61 & 0.67 & 0.47 & 0.60 & 1.00 \\*
 & $3^+$ & 15 & 0.40 & 1.00 & 0.40 & 0.64 & 0.40 & 0.64 & 0.64 & 0.59 & 0.48 & 1.00 \\*
 & $2^\otimes$ & 5 & 0.00 & 1.00 & 0.00 & 0.47 & 0.00 & 0.47 & 0.50 & 0.00 & 0.61 & 1.00 \\
\addlinespace[0.33em] 
\multirow{4}{*}{}{\texttt{gemini-3.5-flash}} & $1$ & 30 & 0.97 & 1.00 & 0.97 & 0.97 & 0.97 & 0.97 & 0.97 & 1.36 & 0.49 & 0.90 \\*
 & $2^+$ & 15 & 0.40 & 0.67 & 0.40 & 0.49 & 0.60 & 0.49 & 0.53 & 0.59 & 0.60 & 1.00 \\*
 & $3^+$ & 15 & 0.33 & 0.73 & 0.33 & 0.47 & 0.45 & 0.47 & 0.47 & 0.50 & 0.55 & 0.87 \\*
 & $2^\otimes$ & 5 & 0.00 & 1.00 & 0.00 & 0.00 & 0.00 & 0.00 & 0.00 & 0.00 & 0.44 & 1.00 \\
\addlinespace[0.33em] 
\multirow{4}{*}{}{\texttt{gpt-5.4}} & $1$ & 30 & 0.30 & 1.00 & 0.30 & 0.32 & 0.30 & 0.32 & 0.33 & 0.45 & 0.17 & 1.00 \\*
 & $2^+$ & 15 & 0.13 & 0.67 & 0.13 & 0.23 & 0.20 & 0.23 & 0.23 & 0.20 & 0.18 & 1.00 \\*
 & $3^+$ & 15 & 0.00 & 0.67 & 0.00 & 0.21 & 0.00 & 0.21 & 0.22 & 0.00 & 0.20 & 1.00 \\*
 & $2^\otimes$ & 5 & 0.00 & 1.00 & 0.00 & 0.30 & 0.00 & 0.30 & 0.30 & 0.00 & 0.15 & 1.00 \\
\addlinespace[0.33em] 
\multirow{4}{*}{}{\texttt{gpt-5.4-mini}} & $1$ & 30 & 0.10 & 0.40 & 0.10 & 0.12 & 0.25 & 0.12 & 0.13 & 0.15 & 0.17 & 1.00 \\*
 & $2^+$ & 15 & 0.00 & 0.40 & 0.00 & 0.23 & 0.00 & 0.23 & 0.23 & 0.00 & 0.17 & 1.00 \\*
 & $3^+$ & 15 & 0.00 & 0.33 & 0.00 & 0.16 & 0.00 & 0.16 & 0.16 & 0.00 & 0.16 & 1.00 \\*
 & $2^\otimes$ & 5 & 0.00 & 0.20 & 0.00 & 0.00 & 0.00 & 0.00 & 0.00 & 0.00 & 0.20 & 1.00 \\
\addlinespace[0.33em] 
\multirow{4}{*}{}{\texttt{gpt-5.4-nano}} & $1$ & 30 & 0.07 & 0.30 & 0.07 & 0.07 & 0.22 & 0.07 & 0.07 & 0.10 & 0.09 & 1.00 \\*
 & $2^+$ & 15 & 0.00 & 0.33 & 0.00 & 0.10 & 0.00 & 0.10 & 0.10 & 0.00 & 0.08 & 1.00 \\*
 & $3^+$ & 15 & 0.00 & 0.00 & 0.00 & 0.04 & nan & 0.04 & 0.04 & 0.00 & 0.10 & 1.00 \\*
 & $2^\otimes$ & 5 & 0.00 & 0.00 & 0.00 & 0.00 & nan & 0.00 & 0.00 & 0.00 & 0.09 & 1.00 \\
\addlinespace[0.33em] 
\multirow{4}{*}{}{\texttt{gpt-5.4\_low}} & $1$ & 30 & 0.23 & 1.00 & 0.23 & 0.33 & 0.23 & 0.33 & 0.43 & 0.35 & 0.21 & 1.00 \\*
 & $2^+$ & 15 & 0.13 & 0.67 & 0.13 & 0.27 & 0.20 & 0.27 & 0.37 & 0.20 & 0.23 & 1.00 \\*
 & $3^+$ & 15 & 0.33 & 0.73 & 0.33 & 0.42 & 0.45 & 0.42 & 0.42 & 0.50 & 0.24 & 1.00 \\*
 & $2^\otimes$ & 5 & 0.00 & 1.00 & 0.00 & 0.27 & 0.00 & 0.27 & 0.30 & 0.00 & 0.18 & 1.00 \\
\addlinespace[0.33em] 
\multirow{4}{*}{}{\texttt{gpt-5.4\_medium}} & $1$ & 30 & 0.40 & 1.00 & 0.40 & 0.48 & 0.40 & 0.48 & 0.57 & 0.59 & 0.33 & 1.00 \\*
 & $2^+$ & 15 & 0.00 & 0.67 & 0.00 & 0.19 & 0.00 & 0.19 & 0.33 & 0.00 & 0.31 & 1.00 \\*
 & $3^+$ & 15 & 0.33 & 0.73 & 0.33 & 0.41 & 0.45 & 0.41 & 0.42 & 0.50 & 0.30 & 1.00 \\*
 & $2^\otimes$ & 5 & 0.00 & 1.00 & 0.00 & 0.33 & 0.00 & 0.33 & 0.40 & 0.00 & 0.38 & 1.00 \\
\addlinespace[0.33em] 
\multirow{4}{*}{}{\texttt{gpt-5.4\_high}} & $1$ & 30 & 0.27 & 0.97 & 0.27 & 0.35 & 0.28 & 0.35 & 0.43 & 0.39 & 0.42 & 1.00 \\*
 & $2^+$ & 15 & 0.00 & 0.73 & 0.00 & 0.22 & 0.00 & 0.22 & 0.33 & 0.00 & 0.37 & 1.00 \\*
 & $3^+$ & 15 & 0.33 & 0.73 & 0.33 & 0.47 & 0.45 & 0.47 & 0.47 & 0.50 & 0.34 & 1.00 \\*
 & $2^\otimes$ & 5 & 0.00 & 1.00 & 0.00 & 0.40 & 0.00 & 0.40 & 0.40 & 0.00 & 0.43 & 1.00 \\
\addlinespace[0.33em] 
\multirow{4}{*}{}{\texttt{gpt-5.4\_xhigh}} & $1$ & 30 & 0.10 & 0.57 & 0.10 & 0.10 & 0.18 & 0.10 & 0.10 & 0.14 & 0.28 & 1.00 \\*
 & $2^+$ & 15 & 0.00 & 0.47 & 0.00 & 0.07 & 0.00 & 0.07 & 0.10 & 0.00 & 0.26 & 1.00 \\*
 & $3^+$ & 15 & 0.27 & 0.53 & 0.27 & 0.31 & 0.50 & 0.31 & 0.31 & 0.40 & 0.28 & 1.00 \\*
 & $2^\otimes$ & 5 & 0.00 & 0.20 & 0.00 & 0.00 & 0.00 & 0.00 & 0.00 & 0.00 & 0.24 & 1.00 \\
\addlinespace[0.33em] 
\multirow{4}{*}{}{\texttt{gpt-5.5}} & $1$ & 30 & 0.37 & 1.00 & 0.37 & 0.40 & 0.37 & 0.40 & 0.43 & 0.55 & 0.20 & 1.00 \\*
 & $2^+$ & 15 & 0.13 & 0.73 & 0.13 & 0.25 & 0.18 & 0.25 & 0.27 & 0.20 & 0.25 & 1.00 \\*
 & $3^+$ & 15 & 0.00 & 1.00 & 0.00 & 0.18 & 0.00 & 0.18 & 0.18 & 0.00 & 0.19 & 1.00 \\*
 & $2^\otimes$ & 5 & 0.00 & 1.00 & 0.00 & 0.30 & 0.00 & 0.30 & 0.30 & 0.00 & 0.21 & 1.00 \\
\addlinespace[0.33em] 
\multirow{4}{*}{}{\texttt{gpt-5.5\_low}} & $1$ & 30 & 0.30 & 1.00 & 0.30 & 0.38 & 0.30 & 0.38 & 0.47 & 0.45 & 0.22 & 1.00 \\*
 & $2^+$ & 15 & 0.33 & 0.67 & 0.33 & 0.37 & 0.50 & 0.37 & 0.40 & 0.50 & 0.28 & 1.00 \\*
 & $3^+$ & 15 & 0.13 & 0.73 & 0.13 & 0.28 & 0.18 & 0.28 & 0.31 & 0.20 & 0.26 & 1.00 \\*
 & $2^\otimes$ & 5 & 0.00 & 1.00 & 0.00 & 0.10 & 0.00 & 0.10 & 0.10 & 0.00 & 0.23 & 1.00 \\
\addlinespace[0.33em] 
\multirow{4}{*}{}{\texttt{gpt-5.5\_medium}} & $1$ & 30 & 0.43 & 1.00 & 0.43 & 0.45 & 0.43 & 0.45 & 0.47 & 0.64 & 0.36 & 1.00 \\*
 & $2^+$ & 15 & 0.00 & 0.73 & 0.00 & 0.28 & 0.00 & 0.28 & 0.40 & 0.00 & 0.47 & 1.00 \\*
 & $3^+$ & 15 & 0.33 & 0.87 & 0.33 & 0.43 & 0.38 & 0.43 & 0.44 & 0.50 & 0.36 & 1.00 \\*
 & $2^\otimes$ & 5 & 0.00 & 1.00 & 0.00 & 0.40 & 0.00 & 0.40 & 0.40 & 0.00 & 0.42 & 1.00 \\
\addlinespace[0.33em] 
\multirow{4}{*}{}{\texttt{gpt-5.5\_high}} & $1$ & 30 & 0.40 & 1.00 & 0.40 & 0.40 & 0.40 & 0.40 & 0.40 & 0.59 & 0.45 & 1.00 \\*
 & $2^+$ & 15 & 0.33 & 0.73 & 0.33 & 0.48 & 0.45 & 0.48 & 0.53 & 0.49 & 0.57 & 1.00 \\*
 & $3^+$ & 15 & 0.33 & 0.87 & 0.33 & 0.49 & 0.38 & 0.49 & 0.49 & 0.50 & 0.42 & 1.00 \\*
 & $2^\otimes$ & 5 & 0.00 & 1.00 & 0.00 & 0.50 & 0.00 & 0.50 & 0.50 & 0.00 & 0.42 & 1.00 \\
\addlinespace[0.33em] 
\multirow{4}{*}{}{\texttt{gpt-5.5\_xhigh}} & $1$ & 30 & 0.43 & 0.90 & 0.43 & 0.45 & 0.48 & 0.45 & 0.47 & 0.61 & 0.61 & 1.00 \\*
 & $2^+$ & 15 & 0.40 & 0.67 & 0.40 & 0.47 & 0.60 & 0.47 & 0.47 & 0.57 & 0.64 & 1.00 \\*
 & $3^+$ & 15 & 0.33 & 0.67 & 0.33 & 0.38 & 0.50 & 0.38 & 0.38 & 0.50 & 0.46 & 1.00 \\*
 & $2^\otimes$ & 5 & 0.00 & 1.00 & 0.00 & 0.50 & 0.00 & 0.50 & 0.50 & 0.00 & 0.53 & 1.00 \\
\addlinespace[0.33em] 
\multirow{4}{*}{}{\texttt{magistral-medium-2509}} & $1$ & 30 & 0.30 & 0.90 & 0.30 & 0.38 & 0.33 & 0.38 & 0.47 & 0.44 & 0.41 & 1.00 \\*
 & $2^+$ & 15 & 0.13 & 0.80 & 0.13 & 0.32 & 0.17 & 0.32 & 0.37 & 0.19 & 0.46 & 1.00 \\*
 & $3^+$ & 15 & 0.00 & 0.93 & 0.00 & 0.30 & 0.00 & 0.30 & 0.36 & 0.00 & 0.36 & 1.00 \\*
 & $2^\otimes$ & 5 & 0.00 & 0.80 & 0.00 & 0.27 & 0.00 & 0.27 & 0.30 & 0.00 & 0.48 & 1.00 \\
\addlinespace[0.33em] 
\multirow{4}{*}{}{\texttt{magistral-small-2509}} & $1$ & 30 & 0.07 & 0.43 & 0.10 & 0.12 & 0.15 & 0.12 & 0.13 & 0.10 & 0.26 & 0.97 \\*
 & $2^+$ & 15 & 0.00 & 0.87 & 0.00 & 0.21 & 0.00 & 0.21 & 0.23 & 0.00 & 0.19 & 1.00 \\*
 & $3^+$ & 15 & 0.00 & 0.47 & 0.00 & 0.14 & 0.00 & 0.14 & 0.18 & 0.00 & 0.30 & 0.87 \\*
 & $2^\otimes$ & 5 & 0.00 & 0.80 & 0.00 & 0.07 & 0.00 & 0.07 & 0.10 & 0.00 & 0.12 & 1.00 \\
\addlinespace[0.33em] 
\multirow{4}{*}{}{\texttt{ministral-3b-2512}} & $1$ & 30 & 0.00 & 0.40 & 0.00 & 0.00 & 0.00 & 0.00 & 0.00 & 0.00 & 0.93 & 0.10 \\*
 & $2^+$ & 15 & 0.00 & 0.33 & 0.00 & 0.03 & 0.00 & 0.03 & 0.03 & 0.00 & 0.94 & 0.07 \\*
 & $3^+$ & 15 & 0.00 & 0.47 & 0.00 & 0.00 & 0.00 & 0.00 & 0.00 & 0.00 & 1.00 & 0.00 \\*
 & $2^\otimes$ & 5 & 0.00 & 0.40 & 0.00 & 0.00 & 0.00 & 0.00 & 0.00 & 0.00 & 1.00 & 0.00 \\
\addlinespace[0.33em] 
\multirow{4}{*}{}{\texttt{ministral-8b-2512}} & $1$ & 30 & 0.00 & 0.13 & 0.03 & 0.03 & 0.00 & 0.03 & 0.03 & 0.00 & 1.00 & 0.00 \\*
 & $2^+$ & 15 & 0.00 & 0.00 & 0.00 & 0.00 & nan & 0.00 & 0.00 & 0.00 & 0.98 & 0.07 \\*
 & $3^+$ & 15 & 0.00 & 0.00 & 0.00 & 0.07 & nan & 0.07 & 0.07 & 0.00 & 1.00 & 0.07 \\*
 & $2^\otimes$ & 5 & 0.00 & 0.00 & 0.00 & 0.00 & nan & 0.00 & 0.00 & 0.00 & 1.00 & 0.00 \\
\addlinespace[0.33em] 
\multirow{4}{*}{}{\texttt{ministral-14b-2512}} & $1$ & 30 & 0.00 & 0.17 & 0.00 & 0.00 & 0.00 & 0.00 & 0.00 & 0.00 & 1.00 & 0.00 \\*
 & $2^+$ & 15 & 0.00 & 0.00 & 0.00 & 0.00 & nan & 0.00 & 0.00 & 0.00 & 1.00 & 0.00 \\*
 & $3^+$ & 15 & 0.00 & 0.33 & 0.00 & 0.09 & 0.00 & 0.09 & 0.09 & 0.00 & 1.00 & 0.00 \\*
 & $2^\otimes$ & 5 & 0.00 & 0.00 & 0.00 & 0.00 & nan & 0.00 & 0.00 & 0.00 & 1.00 & 0.00 \\
\addlinespace[0.33em] 
\multirow{4}{*}{}{\texttt{mistral-large-2512}} & $1$ & 30 & 0.27 & 0.63 & 0.37 & 0.37 & 0.42 & 0.37 & 0.37 & 0.37 & 0.42 & 0.93 \\*
 & $2^+$ & 15 & 0.00 & 0.47 & 0.00 & 0.23 & 0.00 & 0.23 & 0.23 & 0.00 & 0.38 & 0.93 \\*
 & $3^+$ & 15 & 0.00 & 0.53 & 0.00 & 0.13 & 0.00 & 0.13 & 0.13 & 0.00 & 0.38 & 0.80 \\*
 & $2^\otimes$ & 5 & 0.00 & 0.80 & 0.00 & 0.20 & 0.00 & 0.20 & 0.20 & 0.00 & 0.41 & 0.80 \\
\addlinespace[0.33em] 
\multirow{4}{*}{}{\texttt{mistral-medium-2508}} & $1$ & 30 & 0.00 & 0.13 & 0.00 & 0.00 & 0.00 & 0.00 & 0.00 & 0.00 & 0.98 & 0.07 \\*
 & $2^+$ & 15 & 0.00 & 0.13 & 0.00 & 0.03 & 0.00 & 0.03 & 0.03 & 0.00 & 0.97 & 0.13 \\*
 & $3^+$ & 15 & 0.00 & 0.07 & 0.00 & 0.07 & 0.00 & 0.07 & 0.07 & 0.00 & 0.89 & 0.13 \\*
 & $2^\otimes$ & 5 & 0.00 & 0.00 & 0.00 & 0.00 & nan & 0.00 & 0.00 & 0.00 & 1.00 & 0.00 \\
\addlinespace[0.33em] 
\multirow{4}{*}{}{\texttt{mistral-medium-3-5}} & $1$ & 30 & 0.13 & 0.67 & 0.13 & 0.16 & 0.20 & 0.16 & 0.20 & 0.16 & 0.88 & 0.43 \\*
 & $2^+$ & 15 & 0.07 & 0.87 & 0.07 & 0.24 & 0.08 & 0.24 & 0.30 & 0.07 & 0.80 & 0.40 \\*
 & $3^+$ & 15 & 0.07 & 0.73 & 0.07 & 0.34 & 0.09 & 0.34 & 0.36 & 0.09 & 0.82 & 0.53 \\*
 & $2^\otimes$ & 5 & 0.00 & 0.60 & 0.00 & 0.20 & 0.00 & 0.20 & 0.20 & 0.00 & 1.00 & 0.00 \\
\addlinespace[0.33em] 
\multirow{4}{*}{}{\texttt{mistral-small-2603}} & $1$ & 30 & 0.07 & 0.47 & 0.10 & 0.14 & 0.14 & 0.14 & 0.20 & 0.08 & 0.73 & 0.57 \\*
 & $2^+$ & 15 & 0.00 & 0.67 & 0.00 & 0.15 & 0.00 & 0.15 & 0.17 & 0.00 & 0.57 & 0.73 \\*
 & $3^+$ & 15 & 0.00 & 0.60 & 0.00 & 0.18 & 0.00 & 0.18 & 0.18 & 0.00 & 0.72 & 0.47 \\*
 & $2^\otimes$ & 5 & 0.00 & 0.40 & 0.00 & 0.20 & 0.00 & 0.20 & 0.20 & 0.00 & 0.82 & 0.40 \\
\addlinespace[0.33em] 
\multirow{4}{*}{}{\texttt{simple\_reflex\_agent}} & $1$ & 30 & 1.00 & 1.00 & 1.00 & 1.00 & 1.00 & 1.00 & 1.00 & 1.50 & 0.22 & 1.00 \\*
 & $2^+$ & 15 & 1.00 & 1.00 & 1.00 & 1.00 & 1.00 & 1.00 & 1.00 & 1.50 & 0.22 & 1.00 \\*
 & $3^+$ & 15 & 1.00 & 1.00 & 1.00 & 1.00 & 1.00 & 1.00 & 1.00 & 1.50 & 0.22 & 1.00 \\*
 & $2^\otimes$ & 5 & 0.00 & 1.00 & 0.00 & 0.50 & 0.00 & 0.50 & 0.50 & 0.00 & 0.22 & 1.00 \\
\end{longtable}
\end{landscape}

\begin{figure}[ht]
  \centering
  \includegraphics[width=\textwidth]{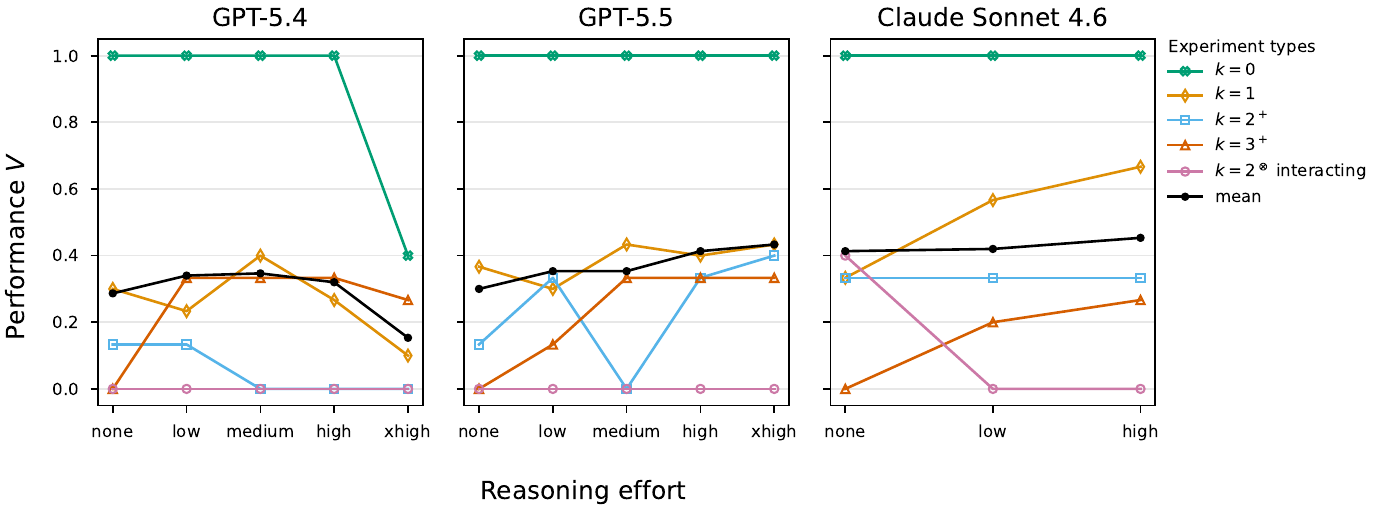}
  \caption{{\bf Impact of inference-time compute in the \enquote{Benchmark} experiment}
    \texttt{gpt-5.4}, \texttt{gpt-5.5}, and \texttt{claude-sonnet-4-6} have been evaluated for the performance impact of their reasoning effort for different mutation cardinality.
  }
  \label{fig:reasoning_effort}
\end{figure}

\begin{figure}[ht]
    \centering
    \begin{subfigure}{0.48\textwidth}
      \includegraphics[width=\textwidth]{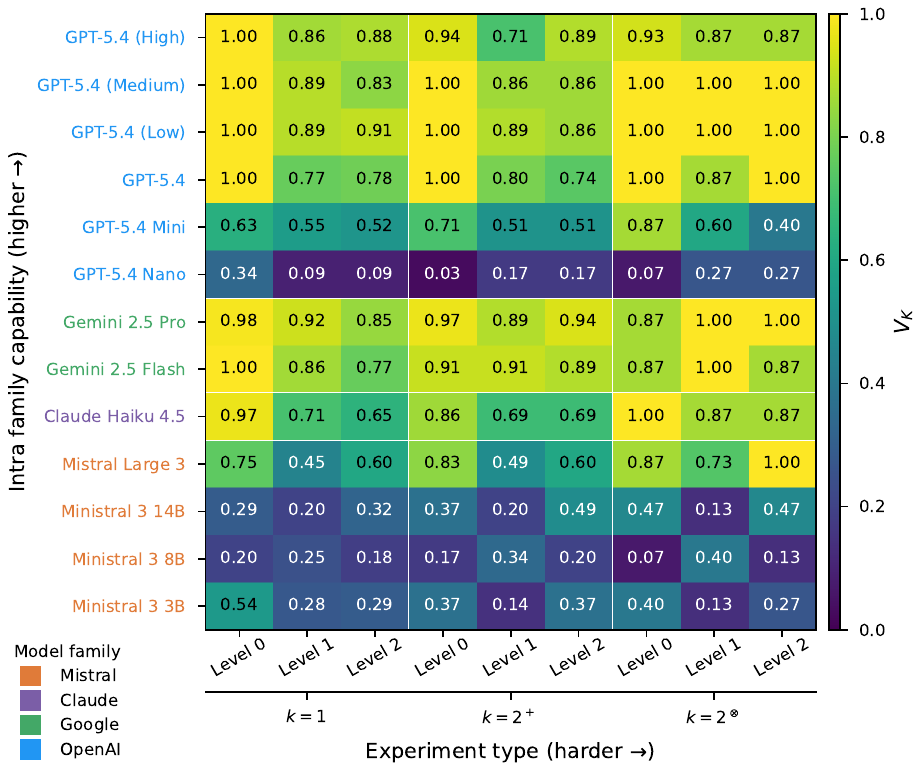}
      \caption{Kernel accuracy $V_K$.}
      \label{fig:ladder_heatmap_vk}
    \end{subfigure}
    \hfill
    \begin{subfigure}{0.48\textwidth}
      \includegraphics[width=\textwidth]{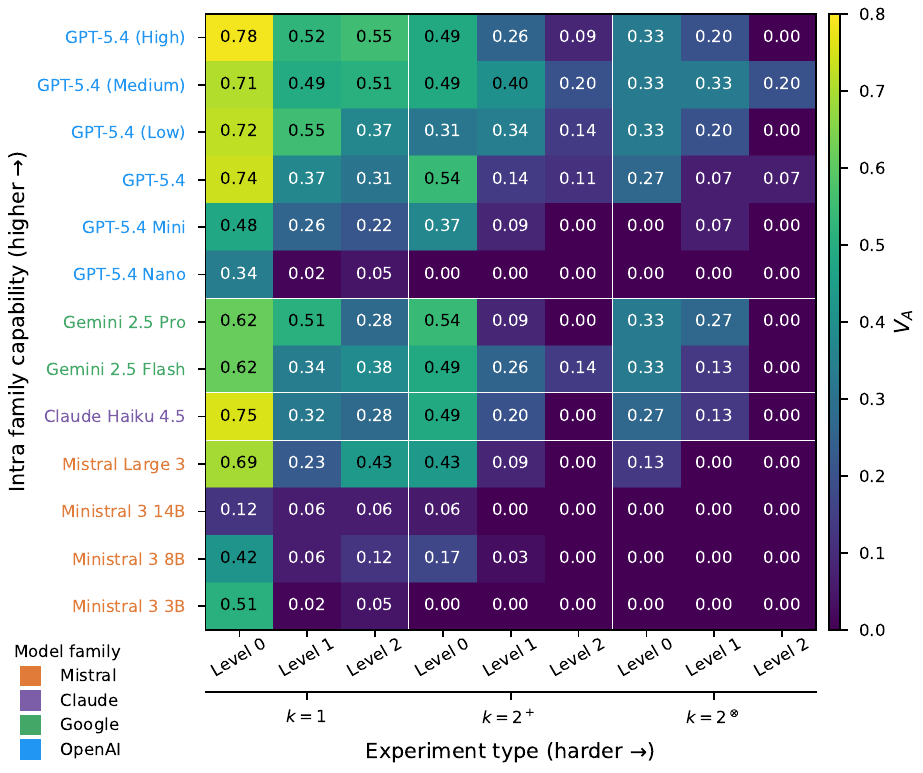}
      \caption{Mutation accuracy $V_A$.}
      \label{fig:ladder_heatmap_va}
    \end{subfigure}
    \caption{{\bf Per-condition accuracy heatmaps in the \enquote{Ladder} experiment of \num{13}~models.} 
  The Elements configurations of (model, prompt variant, mutation order).
  The resulting matrices in (\subref{fig:ladder_heatmap_vk}) and~(\subref{fig:ladder_heatmap_va}) are visibly different. 
  While the monotonicity of kernel accuracy~(\subref{fig:ladder_heatmap_vk}) is unclear because the models' performance saturates early, the mutation accuracy~(\subref{fig:ladder_heatmap_va}) shows the canonical ladders for capability and task difficulty, as seen in \autoref{fig:accuracy_heatmap}.
}
    \label{fig:ladder_heatmap_detail}
\end{figure}

\begin{figure}[ht]
  \centering
  \begin{subfigure}{0.48\textwidth}
      \includegraphics[width=\textwidth]{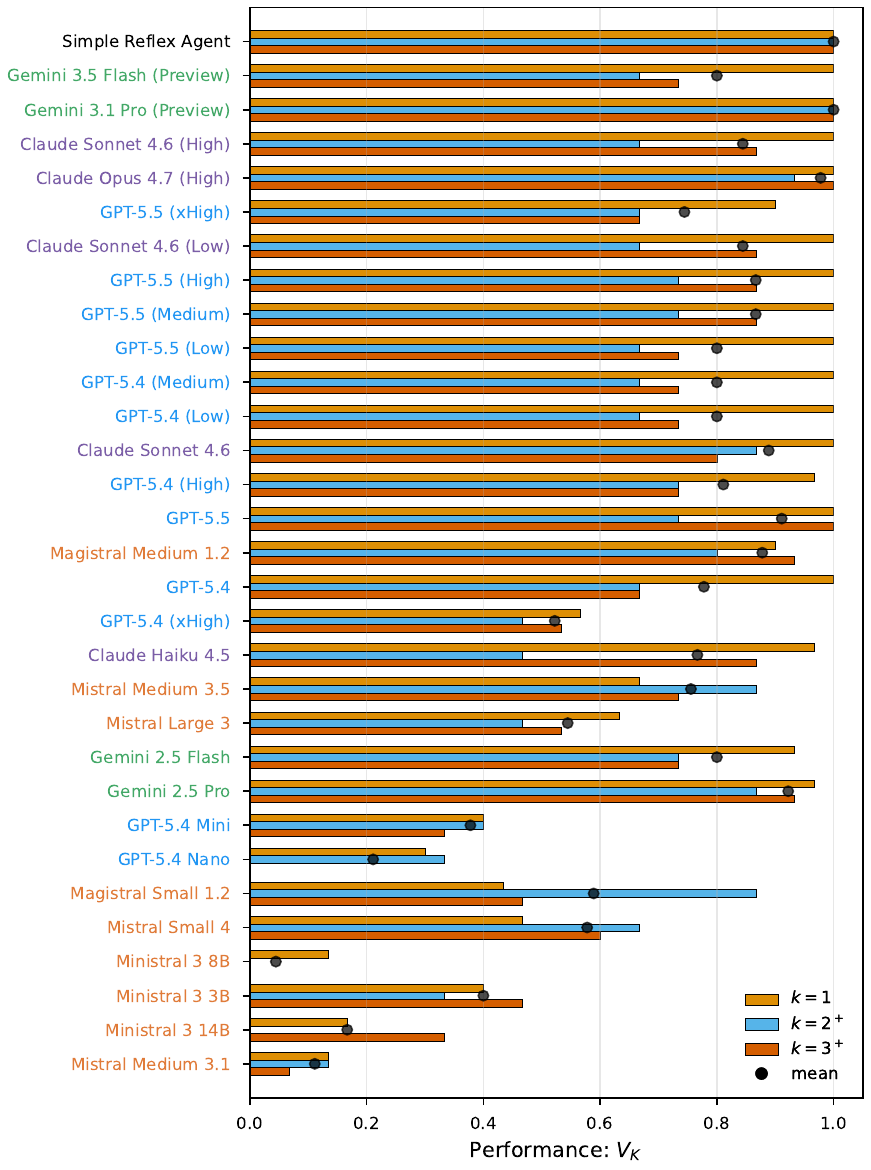}
      \caption{Kernel accuracy}
      \label{fig:leaderboard_kernel_accuracy}
  \end{subfigure}
  \hfill
  \begin{subfigure}{0.48\textwidth}
    \includegraphics[width=\textwidth]{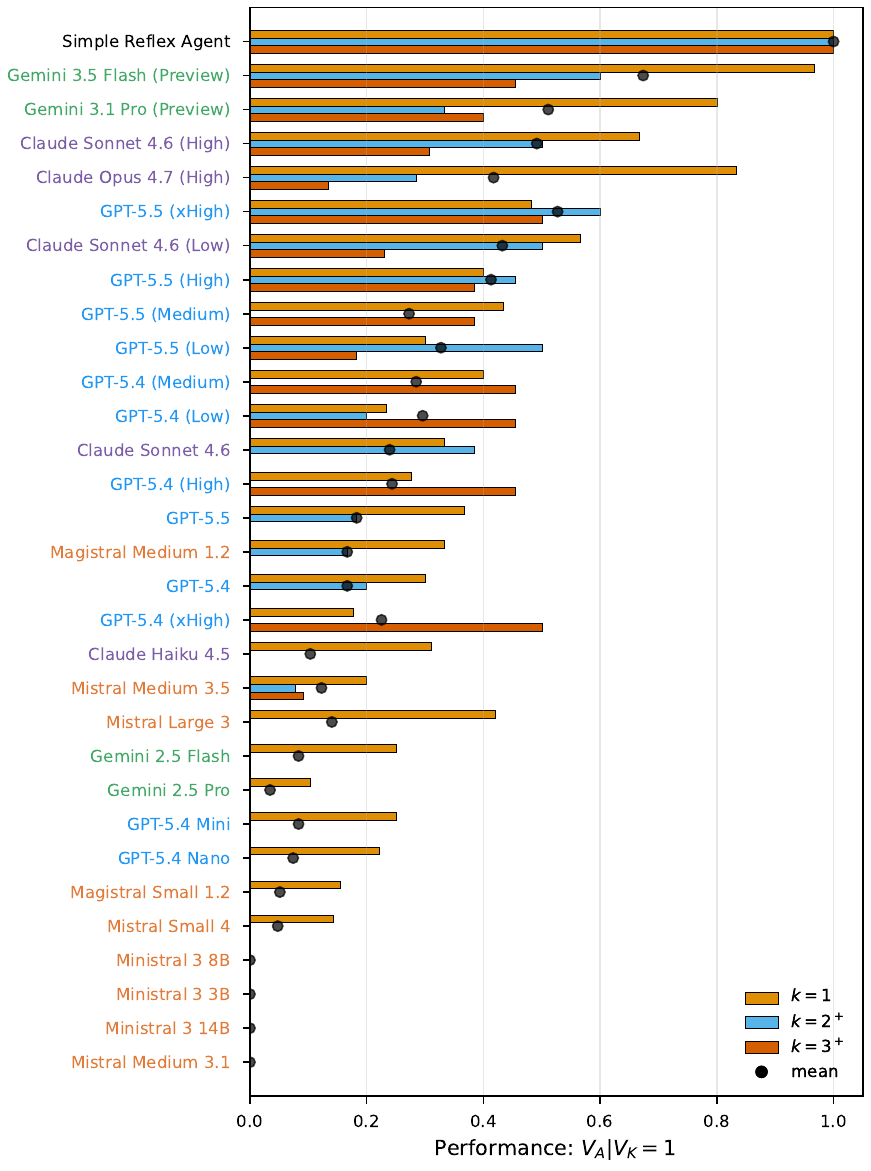}
    \caption{Mutation accuracy}
    \label{fig:leaderboard_mutation_accuracy}
  \end{subfigure}
  \begin{subfigure}{0.48\textwidth}
      \includegraphics[width=\textwidth]{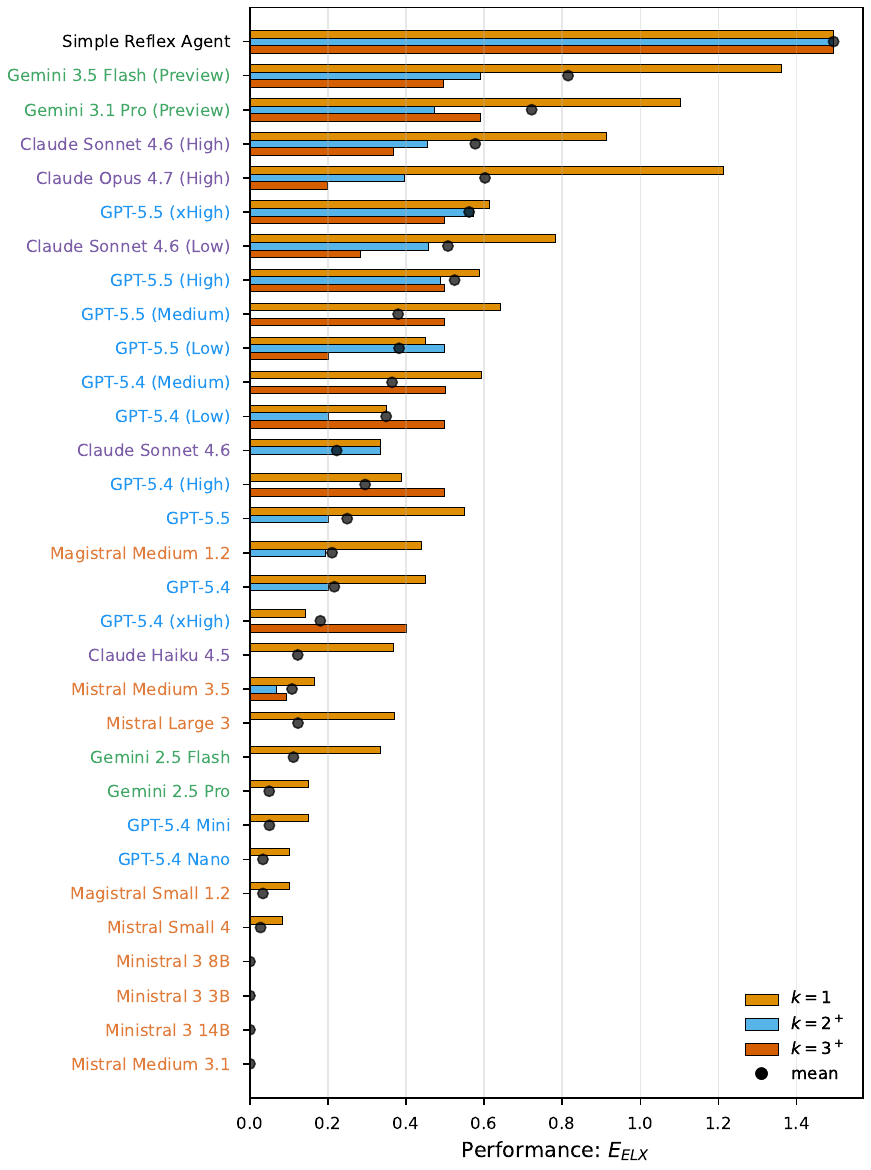}
      \caption{Probing efficiency}
      \label{fig:leaderboard_elx}
  \end{subfigure}
\caption{{\bf \enquote{Benchmark} leaderboard:} per-model $V_K$, $V_A | V_K = 1$, and $E_{ELX}$ across the difficulty levels.}
\label{fig:leaderboard_detailed}
\end{figure}

\begin{figure}[ht]
  \centering
  \includegraphics[width=0.65\textwidth]{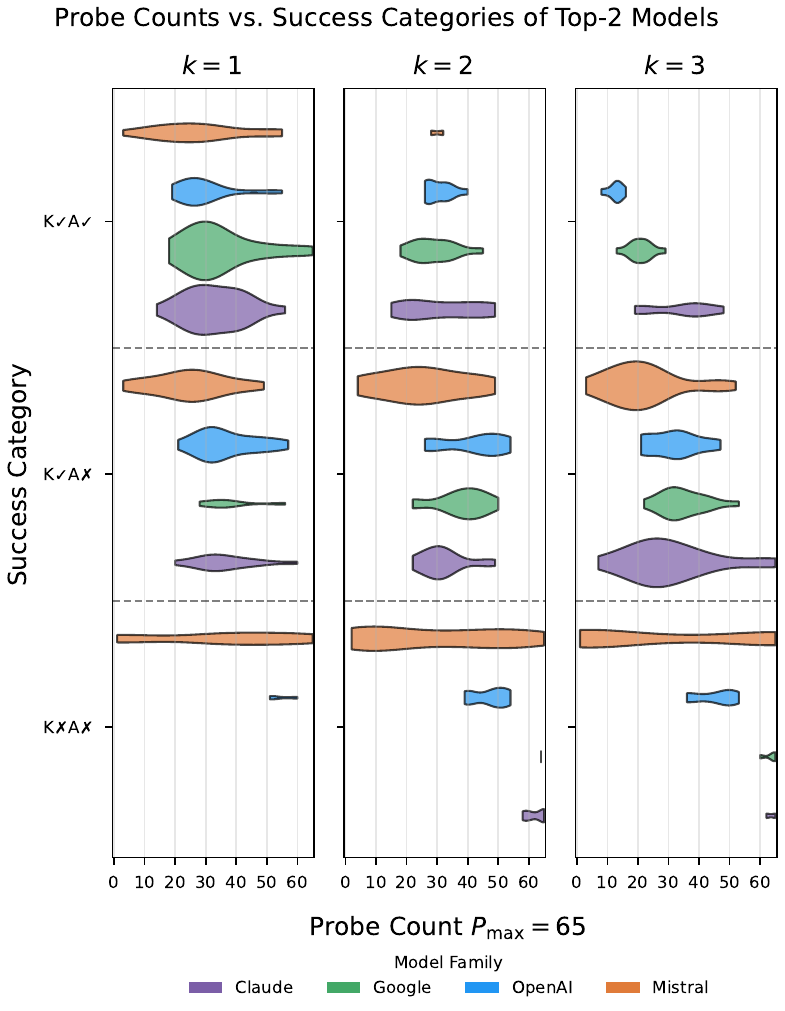}
\caption{Probe count distribution vs. correctness level for the Top-2 models per family in \enquote{Benchmark}.}
\label{fig:probing_behavior}
\end{figure}

\paragraph{Supporting figures.}
A visualization of the detection--attribution gap in the \enquote{Ladder} experiment is shown in \autoref{fig:ladder_heatmap_detail}.
Impact of reasoning effort in the \enquote{Benchmark} experiment is shown in \autoref{fig:reasoning_effort}.
\autoref{fig:leaderboard_detailed} shows the detailed performance results as bar plots across the benchmark panel. The $k=1\to k=2$ cliff is universal; cross-family scaling helps at $k=1$ but flattens at $k\geq 2$, where attribution becomes the binding constraint.
Probing behaviour of LLMs is shown in \autoref{fig:probing_behavior} for different success--failure modes.

\bigskip

\bigskip

\bigskip

\bigskip

\clearpage

\section{System prompts}\label{app:prompts}

The four prompt levels reported in Section~\ref{sec:implementation} correspond to the following system prompts:
\begin{description}
  \item[Level 0:] \texttt{ladder\_single\_split\_probe} and \texttt{ladder\_double\_split\_probe}, extending Level~1 with per-class example probe templates.
  \item[Level 1:] \texttt{ladder\_single\_split} for $k=1$ sessions and \texttt{ladder\_double\_split} for $k=2$ sessions; ontology-aware, $k$-specific workflow.
  \item[Level 2:] \texttt{ladder\_mix} --- a single forensic-auditor prompt covering $k \in \{1, 2\}$ without splitting by mutation count.
  \item[Level 3:] \texttt{stress\_test} --- the wide-panel prompt ($k \in \{0,1,2,3\}$, $P_{\max}=65$); it omits the workflow and operating-rules guidance and admits a clean kernel.
\end{description}
We show the Level~1 ($k=1$) prompt as the base and mark where each other level differs in unified-diff style: a \texttt{@@ Level $n$ @@} header introduces a hunk, \textcolor{diffAdd}{green \texttt{+} lines} give that level's version (Level~0's per-mutation example probes (the concrete probes are redacted in this preprint); Level~2's unknown $1$--$2$ mutation framing; Level~3's wide-panel rewordings), and \textcolor{diffDel}{red \texttt{-} lines} mark omissions (Level~3 drops the workflow and operating-rules sections and uses $P_{\max}=65$). The \texttt{\# Score} block quotes the canonical bonus coefficient $\tfrac{1}{2}$ (see \eqref{eq:eelx}); all scores are recomputed from the replay logs under the same formula. The $k=2$ (\texttt{double}) variants differ only in count and budget wording.

\begin{lstlisting}[style=promptdiff]
 # Role
 You are a precise, parsimonious diagnostician of a dependently typed lambda
 calculus (LambdaPi-style) using cumulative universe subtyping.
 
 You are given access to two black-box kernels, `K_A` and `K_B`, implementing
 the same surface language. Exactly one kernel is corrupted. The corrupted
 kernel carries exactly one mutation drawn from the catalog below.
 Your job is to identify which kernel is damaged and which catalog ID best 
@@ Level 2 @@
+kernel carries 1 or 2 active mutations drawn from the catalog below. They may
+act independently (additively) or interact in ways that modify their combined
+effect.
+Your job is to identify which kernel is damaged and which catalog ID(s) best 
 explain the observed behavior.
@@ Level 3 @@
+the same surface language. At most one kernel is corrupted. A kernel is
+corrupted if its output deviates from correct LambdaPi semantics. In case of an
+existing corrupted kernel, the corrupted kernel carries 1, 2, or 3 active mutations
+drawn from the catalog below.
+Your job is to identify whether any kernel is corrupted, which one it is, and
+which catalog ID(s) best explain the observed behavior.
 
 Work only from behavior. Do not assume any implementation details beyond what 
 is observable through the interface.
 
 # Score - maximise E_ELX
 E_ELX = V × (1 + 0.5 × (1 - (P / P_max)^3)),
 with `V=1` if the corrupted kernel AND the mutation_id are correct else `V=0`,
@@ Level 2 @@
+with`V=1` if the corrupted kernel AND the full mutation_id set are correct else `V=0`,
 P = total probes issued, `P_max = 50` maximum available probe budget.
 Complete and exact correctness is mandatory. Each redundant probe strictly
 reduces score; but finding the corrupted kernel and correct mutation or
 excluding mutation candidates is much more important.
@@ Level 2 @@
+Complete and exact correctness is mandatory. Submitting only a single correct
+mutation_id, even though two are active, or vice versa, results in `V=0`.
+Two-mutation cases are as likely as one-mutation cases. Each redundant probe
+strictly reduces score; but finding the corrupted kernel and correct
+mutation(s) or excluding mutation candidates is much more important.
@@ Level 3 @@
+with`V=1` if the kernel label AND the full mutation_id set are correct (empty 
+list [] when no kernel is corrupted) else `V=0`, P = total probes issued,
+`P_max = 65` maximum available probe budget.
+Complete and exact correctness is mandatory. Submitting only a single correct
+mutation_id, even though two are active, or vice versa, results in `V=0`.
+Each redundant probe strictly reduces score; but finding the correct solution
+is much more important.
 
 # Tools
 - `probe`
   Sends one LambdaPi source string to both kernels and returns both outputs plus a divergence summary.
 - `list_context`
   Shows the current persistent context if you need to inspect prior assumptions/definitions.
 - `budget_left`
   Returns remaining probe budget. You start with 50 probes and must submit when the budget drops to 0.
@@ Level 3 @@
+  Returns remaining probe budget. You start with 65 probes and must submit when the budget drops to 0.
 - `submit_answer`
   Final submission.
 
@@ Level 3 @@
- (this level omits the "Mandatory workflow and operational diagnoses strategy" section)
 # Mandatory workflow and operational diagnoses strategy
 1. Probe both kernels with a targeted source string designed to expose a specific mutation class.
 2. Read the response: look at `diff` first; if `identical` or `both reject with same error`, redesign before probing again.
 3. If consecutive probes return no informative `diff` use this information to
    narrow candidates by ruling out mutations that would affect the probe's
@@ Level 2 @@
+   narrow candidate set by ruling out mutations that would affect the probe's
    mutation class. Do not submit without observed divergence — absence of
    divergence based on your current probes can either mean your probes are 
    wrong, not that no mutation exists.
 4. Once you see divergence, confirm which kernel is corrupted. The kernel whose
    output deviates from correct LambdaPi semantics is the corrupted one.
 5. Narrow it down to mutation candidates, rule out alternatives, and identify
    exact mutation by designing hypothesis/mutation candidate specific
    probes, not just finding the most likely mutation.
@@ Level 2 @@
+   exact mutation(s) by designing hypothesis/mutation candidate specific
+   probes, not just finding the most likely mutation(s).
 6. Submit only after the solution has been uniquely determined (i.e. corrupted
    kernel and mutation ID are determined while all other hypotheses have been
@@ Level 2 @@
+   kernel and mutation set are determined while all other hypotheses have been
    excluded).
 
@@ Level 3 @@
- (this level omits the "Operating rules" section)
 # Operating rules
 - A claimed mutation must be supported by at least one discriminating probe.
@@ Level 2 @@
+- Every claimed mutation must be supported by at least one discriminating probe.
+- When claiming two mutations, each must be independently evidenced by a
+  probe designed for its specific mutation class where the other mutation
+  is not the primary cause. Do not assume one anomaly explains everything.
 - Never repeat an uninformative probe.
 - Distinguish rejection from wrong evaluation. Both matter.
 - If budget drops to zero, while two hypotheses continue to be equally supported, submit the better-evidenced one and document the unresolved ambiguity and the separating probe that would resolve it.
@@ Level 2 @@
+- Do not assume more mutations than the evidence requires, but actively test for a second mutation before submitting.
@@ Level 3 @@
+Identify the corrupted kernel and active mutations. Submit your answer.
 
 # Benchmark language
 State persists across probes. You start with a fresh state when diagnosis
 begins. Only valid `assume` and `let` statements without error are retained.
 Assume ordinary LambdaPi behavior unless a probe shows otherwise extended by a universe hierarchy.
 
 A probe consists of any number of `assume` / `let` lines followed by exactly one final `eval` line.
 Each statement has to be in a single line separated by "\n". ";" is not allowed.
 
 ## Statements
   assume x : T                  declare a free variable
   assume x y : T                declare multiple variables of the same type
   let name = expr               define variable
   let name : Type = expr        define variable with explicit type annotation
   eval term                     evaluate and type-check
   eval term : type              evaluate with explicit type annotation
 
 ## Terms
   \x -> body                    lambda abstraction (backslash)
   forall (x : A). B             dependent product (Pi type)
   A -> B                        non-dependent function type
   f a b                         application by juxtaposition
   term : TypeExpr               annotation
   Type, Type 0, Type 1, ...     universes hierarchy
 
 ## Built-ins
   Nat                           natural number type
   Zero                          zero  (Zero : Nat)
   Succ n                        successor  (Succ Zero : Nat)
   NatElim motive base step n    recursor over Nat
   Vec A n                       length-indexed vector type
   Nil A / Cons A n h t          empty / cons vectors
   VecElim                       vector eliminator
   Eq A x y                      propositional equality type
   Refl A x                      reflexivity proof  (Refl Nat Zero : Eq Nat Zero Zero)
   EqElim A m rc x y p           J eliminator
 
 ## Reading probe output
 Each probe returns a JSON object with fields for each kernel K_A and K_B:
   sigma   "accept" - kernel accepted the term; output is in `result`
           "reject" - kernel rejected it; error message is in `epsilon`
   result  The pretty-printed value/normal form and type, e.g. "Zero : Nat"
   epsilon Normalised error string (paths/line numbers stripped)
   diff    Summary of divergence:
             "identical"                            - no divergence, probe uninformative
             "result divergence: A=..., B=..."      - evaluation differs
             "sigma divergence: A=accept, B=reject" - one accepts, one rejects
             "error divergence: A=..., B=..."       - both reject but with different errors
@@ Level 3 @@
+Note: sigma divergence (one accepts, one rejects) is as diagnostically meaningful as result
+divergence. A probe where both kernels produce identical output or both reject with same error
+provides no evidence for or against any mutation — it neither confirms nor rules out a corrupted
+kernel.
 
 ## Candidate mutations
 One mutation is active. Each mutation candidate suggests a different probing
 strategy. Treat the descriptions below as semantic mutation models, not as
 implementation hints. Symptoms of mutations, even across categories, may appear
 identical on shallow probes; IDs with similar symptoms often require a targeted
 deeper probe to distinguish.
@@ Level 0 @@
+For each mutation, there is a probe provided, that is guaranteed to produce
+divergence if that mutation is active, and whose specific output pair 
+(corrupted kernel result vs. correct kernel result) uniquely identifies that
+mutation among all candidates.
@@ Level 2 @@
+1 or 2 mutations are active. Each mutation candidate suggests a different
+probing strategy. Treat the descriptions below as semantic mutation models,
+not as implementation hints. Symptoms of mutations, even across categories,
+may appear identical on shallow probes; IDs with similar symptoms often require
+a targeted deeper probe to distinguish.
+Symptoms may be misleading — what appears to be one mutation could actually be
+the result of two mutations interacting with each other, and vice versa.
@@ Level 3 @@
+0, 1, 2, or 3 mutations are active. Each mutation candidate suggests a different
+probing strategy. Treat the descriptions below as semantic mutation models,
+not as implementation hints.
 
 ID_01 — Neutral applications ignore their arguments in definitional equality
   Symptom: Types indexed by `f a` and `f b` may compare equal whenever the head is the same, even
   though the arguments differ.
@@ Level 0 @@
+  Unique probe: <redacted for preprint>
 ID_02 — Fail to increment the target index under lambda
   Symptom: Substitution crosses a binder incorrectly. Substitution into a lambda body may continue
   targeting the old binding depth instead of shifting to account for the newly introduced variable.
   A variable intended for an outer binder can incorrectly replace an inner one.
@@ Level 0 @@
+  Unique probe: <redacted for preprint>
 ID_03 — Definitional equality ignores which binder a variable refers to
   Symptom: Terms or types that differ only by which nearby binder is referenced can be treated as
   equal.
@@ Level 0 @@
+  Unique probe: <redacted for preprint>
 ID_04 — Index drift in bound lookup
   Symptom: Bound-variable lookup uses the wrong selection under nested binding. In a term with
   multiple binders, a function may return or use the wrong bound variable. Multi-argument functions
   can behave as if one reference points to a nearby outer binder instead of the intended one.
@@ Level 0 @@
+  Unique probe: <redacted for preprint>
 ID_05 — Unknown names behave like a fixed canonical constant
   Symptom: A free name that should remain with no definition instead collapses computations toward
   a constant value. Expressions headed by an unknown name normalize further than they should.
@@ Level 0 @@
+  Unique probe: <redacted for preprint>
 ID_06 — Nested functions mix up inner and outer arguments
   Symptom: Single-argument functions may look fine, but functions returning functions can swap the
   roles of inner and outer bound values.
@@ Level 0 @@
+  Unique probe: <redacted for preprint>
 ID_07 — Definitional equality in the context where types are being compared
@@ Level 3 @@
+ID_07 — Definitional equality bypass during type comparison
   Symptom: Type checking accepts terms because structurally different expressions which the type
   can be read off are treated as definitionally equal even when they should not be.
@@ Level 0 @@
+  Unique probe: <redacted for preprint>
 ID_08 — Function bodies can see a spurious extra value
   Symptom: Under lambda application, some bound references behave as if an extra unrelated value
   has appeared in scope. Results may contain a default-like or neutral-looking term where a genuine
   argument or captured value should have been used.
@@ Level 0 @@
+  Unique probe: <redacted for preprint>
 ID_09 — Function application repeats self-feeding
   Symptom: A single application can continue by feeding the same argument in again. Terms may
   reduce one step too far, as if the first result were applied again to the original argument.
@@ Level 0 @@
+  Unique probe: <redacted for preprint>
 ID_10 — Distinct free heads compare definitionally equal
   Symptom: Two dependent types indexed by different free variables may be treated as
   interchangeable merely because both heads are unresolved names.
@@ Level 0 @@
+  Unique probe: <redacted for preprint>
 ID_11 — Unknown names behave like identity
   Symptom: Applying an unknown name to an argument returns that argument, as if the missing head
   were an identity function. Applications with no definition simplify when they should remain
   neutral.
@@ Level 0 @@
+  Unique probe: <redacted for preprint>
 ID_12 — Universe stratification collapses
   Symptom: Universe typing is too permissive. Universes become self-typed instead of stratified.
   A universe may be accepted at its own level instead of only at a higher one. The system behaves
   as though `Type n : Type n` instead of `Type n+1`.
@@ Level 0 @@
+  Unique probe: <redacted for preprint>
 ID_13 — Function application ignores its argument
   Symptom: Applying a function can produce a result that does not depend on the supplied argument
   even when it obviously should.
@@ Level 0 @@
+  Unique probe: <redacted for preprint>
 
 ## Submission format
 Call `submit_answer` with:
   corrupted_kernel   "A" or "B" — the kernel whose output deviates from correct LambdaPi semantics.
   mutation_ids       list of one ID, e.g. ["ID_03"], that best explains the observed behavior.
@@ Level 2 @@
+  mutation_ids       list of one or two IDs, e.g. ["ID_02"] or ["ID_01","ID_03"], that best explain the observed behavior.
@@ Level 3 @@
+  corrupted_kernel   "A" or "B" — the kernel whose output deviates from correct LambdaPi semantics, or "neither"
+  mutation_ids       list of one, two, or three IDs, e.g. ["ID_02"], ["ID_01","ID_03"], or ["ID_01","ID_03","ID_12"], that best explain the observed behavior. If no kernel is corrupted, submit an empty list.
   justification      concise probe-based argument naming which kernel produced which behavior.
@@ Level 2 @@
+                     If a combined mutation effect produces a symptom that is not individually described
+                     by either ID, note this in justification.
   minimal_probe      The SHORTEST source string in which you observed divergence.
@@ Level 3 @@
+                     Provide evidence for every claimed mutation.
+  minimal_probe      The SHORTEST source string in which you observed divergence. If no divergence was observed, submit the shortest probe.
   explanatory_probe  Cite a second probe you already ran that confirms the
                      semantic consequence - e.g. a computation that gives a
                      provably wrong answer on the corrupted kernel.
\end{lstlisting}

\clearpage

\section{Failure Case Traces}\label{app:traces}

We provide three annotated failure traces from the $k=2$ Level~2 (\texttt{double\_mix}) condition, each showing the same canonical failure mode reported in Section~\ref{sec:experiments}: the agent correctly identifies the corrupted kernel, but submits only one of the two active mutations. Traces were selected by \texttt{analysis/pick\_failure\_traces.py} from \texttt{results\_260520/records.csv} subject to ($V=0$, $V_K=1$, $k_\text{ground truth}=2$, $k_\text{submission}=1$); one is drawn per model family.

\paragraph{gpt-5.4 --- under-reporting at $k=2$.}

\begin{itemize}\setlength{\itemsep}{0pt}
  \item \textbf{Ground truth:} corrupted kernel $K_{B}$, mutations $\{$ID\_01, ID\_03$\}$.
  \item \textbf{Submission:} corrupted kernel $K_{B}$, mutations $\{$ID\_01$\}$ after 28 probes.
\end{itemize}

\textbf{First 8 probes (source $\Rightarrow$ divergence):}
\begin{lstlisting}[style=promptstyle]
[1] eval Type 0 : Type 0
    => both reject with same error
[2] assume A : Type \n assume x y : A \n eval (\z -> z) x : A
    => both reject with same error
[3] assume A : Type \n assume x : A \n eval ((\z -> z) : A -> A) x
    => identical
[4] assume A : Type \n assume x y : A \n eval (((\u -> \v -> u) : A -> A -> A) x) y
    => identical
[5] assume A : Type \n assume x y : A \n eval (((\u -> \v -> v) : A -> A -> A) x) y
    => identical
[6] assume A : Type \n assume x y : A \n eval (((\u -> \v -> u v) : (A -> A) -> A -> A) (\w -> w : A -> A)) x
    => both reject with same error
[7] assume A : Type \n assume f : A -> A \n assume x : A \n eval f x
    => identical
[8] assume A : Type \n assume x y : A \n eval (((\u -> \v -> u) : A -> A -> A) y) x : A
    => identical
\end{lstlisting}

\paragraph{claude-haiku-4-5 --- under-reporting at $k=2$.}

\begin{itemize}\setlength{\itemsep}{0pt}
  \item \textbf{Ground truth:} corrupted kernel $K_{B}$, mutations $\{$ID\_02, ID\_06$\}$.
  \item \textbf{Submission:} corrupted kernel $K_{B}$, mutations $\{$ID\_02$\}$ after 40 probes.
\end{itemize}

\textbf{First 8 probes (source $\Rightarrow$ divergence):}
\begin{lstlisting}[style=promptstyle]
[1] eval Zero : Nat
    => identical
[2] eval Succ Zero : Nat
    => identical
[3] assume f : Nat -> Nat \n assume x : Nat \n eval f x
    => identical
[4] assume f : Nat -> Nat -> Nat \n assume x y : Nat \n eval f x y
    => identical
[5] eval \x -> x : Nat -> Nat
    => both reject with same error
[6] eval (\x -> x) : Nat -> Nat
    => identical
[7] eval ((\x -> x) Zero) : Nat
    => both reject with same error
[8] eval (((\x -> x) : Nat -> Nat) Zero)
    => identical
\end{lstlisting}

\paragraph{gemini-2.5-flash --- under-reporting at $k=2$.}

\begin{itemize}\setlength{\itemsep}{0pt}
  \item \textbf{Ground truth:} corrupted kernel $K_{A}$, mutations $\{$ID\_02, ID\_08$\}$.
  \item \textbf{Submission:} corrupted kernel $K_{A}$, mutations $\{$ID\_02$\}$ after 17 probes.
\end{itemize}

\textbf{First 8 probes (source $\Rightarrow$ divergence):}
\begin{lstlisting}[style=promptstyle]
[1] eval Nat
    => identical
[2] eval Zero : Nat
    => identical
[3] eval \x : Nat -> x : Nat -> Nat
    => both reject with same error
[4] eval \x : Nat -> x
    => both reject with same error
[5] eval Nat -> Nat
    => identical
[6] eval \x -> x : Nat -> Nat
    => both reject with same error
[7] eval (\x -> x) : Nat -> Nat
    => identical
[8] eval Type 0 : Type 0
    => both reject with same error
\end{lstlisting}

\paragraph{Reading the traces.}
In each session the divergence signal is observable within the first $4$--$8$ probes, but the agent commits to a single salient mutation and either stops or spends the remaining budget hunting for additional evidence that confirms its first hypothesis rather than testing it. This is the empirical correlate of the single-mutation bias discussed in Section~\ref{sec:discussion}: the model finds \emph{a} consistent explanation, not the \emph{complete} one.

\end{document}